\documentclass[11pt]{article}

\usepackage[preprint]{acl}

\usepackage{times}
\usepackage{latexsym}
\usepackage[T1]{fontenc}
\usepackage[utf8]{inputenc}
\usepackage{microtype}
\usepackage[varqu]{inconsolata}
\usepackage{graphicx}
\usepackage{booktabs}
\usepackage{array}
\usepackage{multirow}
\usepackage{amsmath}
\usepackage{xcolor}
\usepackage{xurl}  
\usepackage{enumitem}
\usepackage{pifont}
\newcommand{\cmark}{\ding{51}}

\setlist{nosep,leftmargin=*}

\definecolor{ovBlue}{HTML}{0072B2}
\definecolor{ovGreen}{HTML}{009E73}
\usepackage{upquote}
\usepackage{listings}
\usepackage{tcolorbox}
\tcbuselibrary{skins,breakable,listings}

\definecolor{dkgiPromptFrame}{RGB}{37,55,100}
\definecolor{dkgiPromptBack}{RGB}{246,248,252}
\definecolor{dkgiFileFrame}{RGB}{82,90,102}
\definecolor{dkgiFileBack}{RGB}{248,248,248}
\definecolor{dkgiCaseFrame}{RGB}{36,99,62}
\definecolor{dkgiCaseBack}{RGB}{247,250,247}
\definecolor{dkgiFailFrame}{RGB}{146,40,36}
\definecolor{dkgiFailBack}{RGB}{252,247,246}
\definecolor{dkgiOutPol}{RGB}{24,82,160}
\definecolor{dkgiToolHint}{RGB}{176,74,18}
\definecolor{dkgiSeen}{RGB}{78,92,118}

\newcommand{\dkgimark}[2]{\tcbox[on line,boxsep=0pt,left=1.5pt,right=1.5pt,top=0.7pt,bottom=0.4pt,arc=1pt,
  boxrule=0pt,colback=#1,coltext=white,fontupper=\sffamily\bfseries\fontsize{6.5}{7}\selectfont]{#2}}
\newcommand{\outpolmark}{\dkgimark{dkgiOutPol}{R}}
\newcommand{\toolhintmark}{\dkgimark{dkgiToolHint}{T}}

\colorlet{dkgiWrap}{black!45}
\newcommand{\dkgiwrap}{\makebox[1em][l]{\textcolor{dkgiWrap}{\scalebox{0.8}{$\hookrightarrow$}}}}
\newcommand{\dkgiwrapmark}{\textcolor{dkgiWrap}{$\hookrightarrow$}}

\lstdefinestyle{dkgibase}{columns=fullflexible,keepspaces=true,upquote=true,showstringspaces=false,
  aboveskip=0pt,belowskip=0pt,extendedchars=false,tabsize=8,
  breaklines=true,breakatwhitespace=true,breakautoindent=false,prebreak={},postbreak={}}
\lstdefinestyle{dkgiprompt}{style=dkgibase,basicstyle=\ttfamily\fontsize{8}{9.6}\selectfont,breakindent=1.5em,
  moredelim=**[is][\color{dkgiOutPol}\outpolmark\,]{@<}{>@},
  moredelim=**[is][\color{dkgiToolHint}\toolhintmark\,]{@|}{|@}}
\lstdefinestyle{dkgicase}{style=dkgibase,basicstyle=\ttfamily\fontsize{7.5}{9}\selectfont,breakindent=0pt,
  postbreak=\mbox{\hspace{2.5em}\dkgiwrap},literate={~>}{{\dkgiwrap}}2 {\\\$}{{\textdollar}}1,
  moredelim=**[is][\bfseries]{@<}{>@},
  moredelim=**[is][\color{dkgiSeen}]{@|}{|@}}

\tcbset{dkgiframe/.style={enhanced jigsaw,boxsep=0pt,
  boxrule=0.6pt,arc=1.2pt,left=4.5pt,right=4.5pt,top=4pt,bottom=4pt,
  toptitle=2.2pt,bottomtitle=2pt,lefttitle=4.5pt,
  coltitle=white,fonttitle=\bfseries\fontsize{8.5}{10}\selectfont,
  before skip=7pt plus 2pt,after skip=7pt plus 2pt},
  dkgibox/.style={dkgiframe,listing only,lines before break=4}}

\newtcblisting{dkgiprompt}[2][]{dkgibox,colframe=dkgiPromptFrame,colback=dkgiPromptBack,
  title={#2},title after break={#2 (continued)},listing options={style=dkgiprompt},#1}
\newtcblisting{dkgifile}[2][]{dkgibox,colframe=dkgiFileFrame,colback=dkgiFileBack,
  title={#2},title after break={#2 (continued)},listing options={style=dkgiprompt},#1}
\newtcblisting{dkgicase}[2][]{dkgibox,colframe=dkgiCaseFrame,colback=dkgiCaseBack,left=3pt,right=3pt,
  title={#2},title after break={#2 (continued)},listing options={style=dkgicase},#1}
\newtcblisting{dkgicase*}[2][]{dkgibox,colframe=dkgiFailFrame,colback=dkgiFailBack,left=3pt,right=3pt,
  title={#2},title after break={#2 (continued)},listing options={style=dkgicase},#1}
\newtcolorbox{dkgiquery}[1]{dkgiframe,colframe=dkgiFileFrame,colback=white,title={#1},
  fontupper=\small,before upper={\setlength{\parskip}{2pt}}}

\newcommand{\dkgiph}[1]{\texttt{<#1>}}

\newcommand{\dkgi}{DKGI}
\newcommand{\hone}{H@1}
\newcommand{\rtwenty}{R@20}
\newcommand{\vfull}{Full \dkgi{}}
\newcommand{\vshell}{Shell only}
\newcommand{\vtoolscode}{Graph tools + code}
\newcommand{\vtools}{Graph tools}
\newcommand{\vretr}{Retrieval only}
\newcommand{\vnoedges}{No edges}
\newcommand{\vcodex}{\vfull{} (Codex)}
\newcommand{\vone}{One program}

\newsavebox{\promptsavebox}
  {\par\smallskip\noindent\begin{lrbox}{\promptsavebox}%
   \begin{minipage}{\dimexpr\linewidth-2\fboxsep-2\fboxrule\relax}%
   \footnotesize\ttfamily\raggedright\setlength{\parskip}{0.45\baselineskip}}%
  {\end{minipage}\end{lrbox}\fbox{\usebox{\promptsavebox}}\par\smallskip}

\newcommand{\qq}{\textquotedbl{}}

\title{Knowledge Graphs as Files:\\
An Agent Harness Where Tool Creation and Tool Use Co-Evolve}

\author{
    Yuyang Bai\textsuperscript{\rm 1}, Yu Wang\textsuperscript{\rm 2}, Yu Zhang\textsuperscript{\rm 1} \\
    \textsuperscript{\rm 1}Texas A\&M University,
    \textsuperscript{\rm 2}University of Georgia \\
    \texttt{\{ybai, yuzhang\}@tamu.edu}
}

\begin{document}
\maketitle

\begin{abstract}
Searching text-rich knowledge graphs requires following relations while examining textual evidence.
Predefined graph interfaces constrain what evidence agents can access, even when they write code to combine tool calls or process retrieved subgraphs.
We propose direct knowledge graph interaction (\dkgi{}), a harness in which language model agents create graph access tools directly over complete node and edge files.
The agent writes Python code to follow relations, combine node sets, and select textual evidence, interleaving execution with shell search and file inspection.
This lets agents process large node sets in bulk while returning selected text for the model to assess.
Execution feedback guides tool creation and revision, while the evolving tools determine what evidence the agent gathers next.
On benchmark datasets spanning three domains, training-free \dkgi{} exceeds the strongest published baseline by at least 6.7 Hit@1 points on every graph, even with a minimal harness.
Controlled ablations show the benefits of multi-turn programming over shell-only access and predefined graph tools, while trajectories show agents revising code to extend searches or address errors.
These findings highlight the value of allowing tool creation and tool use to co-evolve during graph search.
Our code and data are available at \url{https://github.com/leopoldwhite/DKGI}.
\end{abstract}

\section{Introduction}
\label{sec:intro}
\begin{figure*}[t]
\centering
\includegraphics[width=\textwidth]{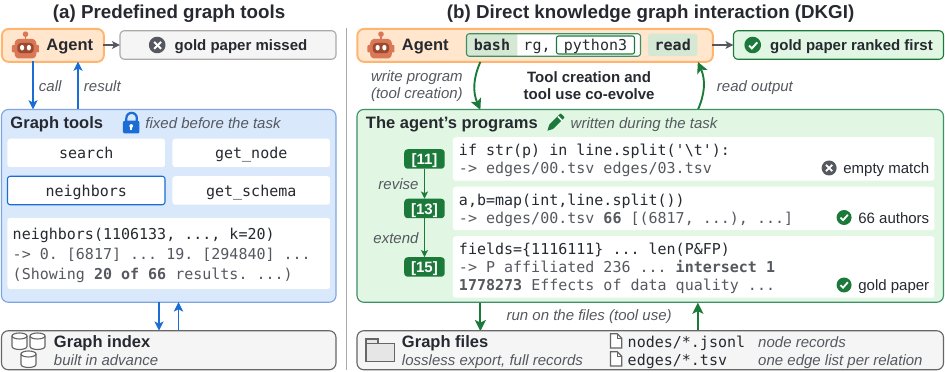}
\caption{An example from the STaRK-MAG dataset~\citep{wu2024stark}: \emph{``Papers on data quality enhancement algorithms affiliated with the University of the Balearic Islands''}. (a) A \texttt{neighbors} call can only return 20 (out of 66) researchers affiliated with the university. The agent never encounters the gold paper in this run. (b) In our \dkgi{}, tool creation and tool use \textbf{co-evolve}: execution feedback guides how the agent revises and extends its code, and the revised code gathers further evidence, ultimately isolating the gold paper. Bracketed numbers denote call indices.}
\label{fig:overview}
\vspace{-1em}
\end{figure*}

\begin{table}[t]
\centering
\fontsize{8}{9.2}\selectfont
\definecolor{dkgiGreen}{HTML}{1B7A3A}
\definecolor{toolBlue}{HTML}{1A6DD4}
\definecolor{inkTwo}{HTML}{55595F}
\newcommand{\pmark}{\textcolor{toolBlue}{\cmark}}
\newcommand{\amark}{\textcolor{dkgiGreen}{\cmark}}
\newcommand{\smark}{\textcolor{dkgiGreen}{(\cmark)}}
\newcommand{\nmark}{\textcolor{inkTwo!55}{--}}
\newcommand{\hd}[2]{\begin{tabular}[b]{@{}c@{}}\fontsize{7.5}{8.4}\selectfont#1\\\fontsize{7.5}{8.4}\selectfont#2\end{tabular}}
\newcommand{\hs}[1]{{\fontsize{7.5}{8.4}\selectfont#1}}
\newcommand{\ex}[1]{\multicolumn{5}{@{}l@{}}{\hspace{0.7em}\fontsize{7.5}{8.6}\selectfont\color{inkTwo}#1}}
\setlength{\tabcolsep}{0pt}\setlength{\aboverulesep}{1.2pt}\setlength{\belowrulesep}{2pt}
\begin{tabular*}{\columnwidth}{@{\extracolsep{\fill}}>{\raggedright\arraybackslash}p{1.21in}*{4}{>{\centering\arraybackslash}p{0.455in}}@{}}
\toprule
& \multicolumn{1}{c}{\makebox[0pt][c]{\textcolor{toolBlue}{Predefined}}} & \multicolumn{3}{c}{\textcolor{dkgiGreen}{Written by the agent}} \\
\cmidrule(lr){2-2}\cmidrule(l){3-5}
Setting & \hd{graph API}{/ retriever} & \hs{code} & \hs{revisions} & \hd{direct file}{access} \\
\midrule
\ding{172} Direct tool use & \pmark & \nmark & \nmark & \nmark \\
\ex{ToG, Graph-CoT, ARK, GRASP} \\
\ex{\emph{\vtools{}}} \\
\midrule
\ding{173} Code over a graph interface & \pmark & \amark & \smark & \nmark \\
\ex{semantic parsing, KG-Agent, Code-on-Graph, AvaTaR} \\
\ex{\emph{\vtoolscode{}}} \\
\midrule
\textbf{\textcolor{dkgiGreen}{\ding{174} Direct knowledge graph interaction}} & \nmark & \amark & \amark & \amark \\
\ex{\emph{\vfull{}}} \\
\bottomrule
\end{tabular*}
\caption{Graph access and agent-written code across three settings. Direct file access lets agent-written code access all node and edge files without a predefined graph API or subgraph retriever. ``\textcolor{dkgiGreen}{(\cmark)}'' indicates support in some systems. Italicized entries denote our controlled variants (\S\ref{sec:ablation}).}
\label{tab:taxonomy}
\vspace{-1.5em}
\end{table}

Knowledge graphs (KGs) store entities and typed relations~\citep{pan2024unifying}. In text-rich KGs, also known as semi-structured knowledge bases, nodes carry additional text information such as product pages, paper abstracts, or drug descriptions~\citep{wu2024stark}. Requests over these graphs often combine relational and textual constraints. For example, to find \emph{``papers by co-authors of X on topic Y''}, a large language model (LLM) agent must locate X, identify the other authors of X's papers, collect their papers, and check which ones address Y. The challenge is therefore to connect evidence across records while inspecting the text that supports each candidate.

Existing methods typically access the graph through specialized interfaces. These interfaces include database query languages~\citep{li2023kbbinder,gu2023pangu,luo2024chatkbqa}, retrievers that return reasoning paths or subgraphs~\citep{luo2024rog,he2024gretriever}, and tools for entity search, node lookup, and neighbor expansion~\citep{sun2024tog,jin2024graphcot,polonuer2026ark}. Beyond direct tool use, some agents write code that combines predefined graph tool calls~\citep{wu2024avatar,jiang2024kgagent}, while others generate code to traverse and process graph objects provided by a dedicated subgraph retriever~\citep{ding2026codeongraph}. Table~\ref{tab:taxonomy} contrasts direct tool use (Setting \ding{172}) with these forms of code-based interaction (Setting \ding{173}). Across these designs, agents have different degrees of freedom to compute over the available data. Nevertheless, \textbf{access to the underlying graph remains governed by a predefined interface}.
Such interfaces determine how much evidence each operation exposes and how the agent can inspect it. Figure~\ref{fig:overview}(a) illustrates this on a request for ``\textit{papers on data quality enhancement algorithms affiliated with the University of the Balearic Islands}''. A \texttt{neighbors} call returns only some of the institution's affiliated researchers. The agent must choose which researchers' papers to examine and make calls to check whether those papers satisfy the topic constraint. In the illustrated run, it never encounters the gold paper. Writing code to batch these calls can simplify this process, but each call still inherits the API's limits on the evidence it returns.

To bridge this gap, we propose \emph{direct knowledge graph interaction} (\dkgi{}), a harness in which agents write and revise their own code to access text-rich KGs. Inspired by the idea of direct corpus interaction over text corpora~\citep{li2026beyond}, we expose both textual node records and typed graph relations as files, together with the graph schema. Unlike Setting \ding{173} in Table~\ref{tab:taxonomy}, where code uses graph query languages or APIs, or operates on retrieved subgraphs, \dkgi{} \textbf{lets the agent implement graph access directly over the complete node and edge files}. The agent writes Python code to follow relations, combine sets of nodes, and inspect textual attributes. It freely interleaves code execution with shell search and record inspection, refining its code as new evidence becomes available. We provide the graph files, their format description, and an execution environment. The agent constructs the graph access tools that each request needs (Setting \ding{174} in Table~\ref{tab:taxonomy}).

On the same request, \dkgi{} writes and revises graph access code (\emph{tool creation}) and executes it to gather evidence (\emph{tool use}). Execution feedback guides which access operations to create or refine. The resulting code, in turn, changes how the agent explores the graph and what evidence it obtains. Tool creation and tool use thus \textbf{co-evolve}: the agent develops its tools as it searches, and its evolving tools guide the subsequent search (Figure~\ref{fig:overview}(b)).

We evaluate \dkgi{} as a training-free method on the three KGs (Amazon, MAG, and Prime) from the STaRK benchmark~\citep{wu2024stark}, using both the minimal Pi harness~\citep{zechner2026pi} and the more feature-rich Codex CLI~\citep{openai2026codexcli}. Even in Pi, our best \dkgi{} configuration exceeds the strongest published baseline~\citep{tao2026grasp} by at least 6.7 Hit@1 points on every dataset. Controlled ablations examine the role of agent-written graph access code. With the backbone, harness, and budgets held fixed, shell-only access trails predefined graph tools, whereas allowing the agent to write and execute Python code improves overall Hit@1 by 6.9 points and reverses this comparison.
A single Python execution recovers little of this gain, supporting the value of multi-turn interaction. Analysis of trajectories further shows that agents use execution feedback to create and refine graph access code, which in turn guides further search. Tool creation and tool use thus co-evolve during graph search.

To summarize, our contributions are:
\begin{itemize}
\item We introduce \dkgi{}, a harness in which agents create and refine graph access tools directly over complete node and edge files.
\item We show that training-free \dkgi{} achieves state-of-the-art performance on the STaRK benchmark using a minimal agent harness.
\item Controlled ablations and trajectory analyses show the benefits of interleaving tool creation and tool use and characterize how the two co-evolve during agent--KG interactions.
\end{itemize}

\section{Direct Knowledge Graph Interaction}
\label{sec:method}

\subsection{Overview and Notation}
\label{sec:task}
\dkgi{} places complete graph files in an agent harness. The harness provides an execution environment and returns observations, while the agent determines how to acquire and combine the evidence needed for each request (Figure~\ref{fig:overview}).

Formally, a text-rich KG is a graph $\mathcal{G}=(\mathcal{V},\mathcal{E})$ with typed nodes and relations, in which each node $v$ carries attributes $x_v$, including text~\citep{wu2024stark}. Given a query $q$ that may combine relational and textual constraints, the agent returns candidate nodes ranked by their relevance to $q$. We adopt a ReAct-style interaction loop~\citep{yao2023react}, in which reasoning over the query and previous observations guides subsequent actions. At step $t$, the LLM agent $\pi$ selects an action $a_t$ and receives an observation $o_t$,
\[
a_t\sim\pi(\cdot\mid q,a_{<t},o_{<t}),\quad o_t=\mathrm{view}(\mathrm{run}(a_t)),
\]
where $\mathrm{run}(\cdot)$ executes the action in an environment containing the graph files and the task's scratch files, and $\mathrm{view}(\cdot)$ determines what the agent observes. The agent uses shell tools for file inspection (e.g., \texttt{cat}, \texttt{head}, \texttt{sed}) and text search (e.g., \texttt{grep}, \texttt{rg})~\citep{li2026beyond}. More importantly, it writes and executes Python code to implement graph access directly over the node and edge files, following relations, intersecting node sets, and filtering textual attributes. Writing or revising this access code constitutes \emph{tool creation}, while executing it constitutes \emph{tool use}. We next describe the graph representation (\S\ref{sec:files}), agent-written graph operations (\S\ref{sec:ops}), output handling (\S\ref{sec:outputs}), and how tool creation and tool use co-evolve (\S\ref{sec:toolbox}).

\subsection{Knowledge Graphs as Files}
\label{sec:files}
Before answering queries, \dkgi{} exports a KG into text files $\Phi(\mathcal{G})=(\mathcal{N},\{\mathcal{E}_r\}_{r\in \mathcal{R}},\mathcal{S})$, where $\mathcal{R}$ is the set of relation types. The node data $\mathcal{N}$ are stored as one JSON line per node $v$, containing its ID, type, and attributes $x_v$. Each relation type $r$ has a tab-separated edge list $\mathcal{E}_r\subseteq \mathcal{V}\times \mathcal{V}$ whose columns identify the source and target nodes. The schema $\mathcal{S}$ describes the node and relation types. This representation preserves all nodes, attributes, and edges, so each query has access to the same complete graph.

The file layout supports both text search and code execution. Text search locates nodes by matching their attributes, while Python can extract the fields needed for further reasoning. A format description specifies the file locations and how to locate nodes by ID. Edge endpoints let code follow a relation in either direction, with the schema guiding interpretation of entity roles. The agent is directed to inspect the format description and schema at the start of a task, then decides which nodes, relations, and attributes to inspect. Appendix~\ref{app:corpus} documents the concrete export layout used in our experiments.

\subsection{Graph Operations Written by the Agent}
\label{sec:ops}
The agent writes Python code that reads $\Phi(\mathcal{G})$ directly and implements the access procedure needed for the current request. To give some examples, we group the operations observed in the agent's code from our experiments into six categories. They also form the basis of our trajectory analysis (\S\ref{sec:analysis}). Note that these categories describe \textbf{generated code} (i.e., \textbf{created tools}), rather than predefined functions supplied by the harness.
\begin{itemize}[leftmargin=*,itemsep=1pt,topsep=2pt]
  \item A \emph{full node scan} applies a selection $\sigma_\phi(\mathcal{N})$ to select nodes whose attributes satisfy a condition $\phi$, extracting the relevant fields.
  \item A \emph{lookup} retrieves a node by its ID. Looking up or following IDs from earlier outputs constitutes a \emph{check of earlier IDs}.
  \item A \emph{one-hop traversal} computes $\mathsf{N}_r(\mathcal{U})=\{v'\mid (v,v')\in \mathcal{E}_r,\ v\in \mathcal{U}\}$ for a set of nodes $\mathcal{U}$, or $\mathsf{N}_r^{-1}(\mathcal{U})$ in the reverse direction.
  \item A \emph{multi-hop traversal} composes one-hop traversals, $\mathsf{N}_{r_k}\circ\dots\circ \mathsf{N}_{r_1}$.
  \item \emph{Set operations} compute unions, intersections, or differences of ID sets.
  \item \emph{Scoring or ranking} sorts candidates, selects the top ones, or accumulates scores or counts.
\end{itemize}
The agent can compose these operations in Python code. For example, for the request \emph{``papers by co-authors of X on topic Y''}, if we let $w$ denote $\xrightarrow{\textit{writes}}$ and $h$ denote $\xrightarrow{\textit{has topic}}$, then code can compute
\[
\mathsf{N}_w\!\left(\mathsf{N}_w^{-1}(\mathsf{N}_w(\{X\}))\setminus\{X\}\right)\cap \mathsf{N}_h^{-1}(\{Y\})
\]
to obtain candidate paper IDs. The code can use these IDs to retrieve abstracts from the node files. The returned text helps the agent assess relevance and decide whether further graph traversal or text inspection is needed. Appendix~\ref{app:cases} illustrates such combinations in concrete searches.

Section~\ref{sec:ablation} compares this design with using predefined graph tools directly or through Python code under controlled conditions.

\subsection{Large Outputs as Files}
\label{sec:outputs}
Broad searches and graph traversals can produce outputs that exceed the model's context budget. Output handling therefore affects both the current observation and the evidence available for later actions~\citep{sen2026grepneedagentharnesses}. Following the use of output files in Pi~\citep{zechner2026pi} and DR-DCI~\citep{lu2026drdci}, our Pi implementation saves the complete output $y_t=\mathrm{run}(a_t)$ of every shell command as a file $\ell_t$ in the task's scratch directory. It presents the model with
\[
\mathrm{view}(y_t)=\begin{cases} y_t, & |y_t|\le B,\\ \mathrm{head}(y_t)\Vert n_t\Vert\mathrm{tail}(y_t), & |y_t|>B, \end{cases}
\]
where $B$ is the display threshold, $\mathrm{head}(\cdot)$ and $\mathrm{tail}(\cdot)$ return the beginning and end of $y_t$, and the notice $n_t$ gives its full size and the path $\ell_t$. This \emph{head-and-tail truncation} bounds what enters the model's context while retaining the complete execution result. Later code can search or filter $\ell_t$, recover IDs omitted from the displayed excerpt, or combine its contents with other evidence. Access code can also return compact candidate lists or selected text fields, giving the agent control over the evidence presented in subsequent observations. Thus an output of tool use can become an input to subsequent tool creation. Appendix~\ref{app:harness} specifies the thresholds and notice, while \S\ref{sec:implementation} describes the differences between harnesses.

\subsection{Co-Evolving Toolbox}
\label{sec:toolbox}
Execution provides candidate IDs, intermediate sets, empty matches, and error messages. The agent can use these observations in later turns to choose another traversal, narrow a filter, inspect a file's structure, or revise earlier code. The revised code determines which nodes and relations subsequent executions examine. Tool creation and tool use therefore \textbf{co-evolve}: evidence guides the construction of access procedures, and those procedures guide further evidence gathering.

In our Pi implementation, the toolbox $\mathcal{T}_t$ consists of access code retained in the interaction history for reuse and revision, while $\mathcal{L}_t$ denotes the saved shell-output files. When an action executes code $p_t$ and saves an output $\ell_t$, these sets become
\[
\mathcal{T}_{t+1}=\mathcal{T}_t\cup\{p_t\},\qquad \mathcal{L}_{t+1}=\mathcal{L}_t\cup\{\ell_t\}.
\]
For actions that contribute no access code or no saved output, the corresponding set is unchanged. Subsequent code can reuse snippets from $\mathcal{T}_t$, read files in $\mathcal{L}_t$, and incorporate IDs from earlier observations. It can revise a failed computation or extend an earlier traversal with new constraints. The agent can also return to shell search or file inspection whenever it needs more evidence. This process develops the access code within a task, without model training or a tool library learned across tasks. Section~\ref{sec:analysis} examines these behaviors, with the measurement rules in Appendix~\ref{app:behaviour}.

\subsection{Implementation}
\label{sec:implementation}
We instantiate \dkgi{} in two general-purpose harnesses: the minimal Pi~\citep{zechner2026pi} and the more feature-rich Codex CLI~\citep{openai2026codexcli}. Pi provides a simple setting for controlled ablations of graph access and agent-written code, with the model, harness, and budgets held fixed (\S\ref{sec:ablation}). Codex tests the same design with richer runtime support. Both provide read-only graph files, a format description, a schema, and a writable scratch directory. Following \citet{li2026beyond}, the Pi implementation supplements its \texttt{bash} and \texttt{read} tools with lightweight context management that retains code while replacing older tool results with placeholders as the context fills. We add the graph export, task prompt, output handling of \S\ref{sec:outputs}, and an execution sandbox. In Codex, the model invokes shell commands, including Python, through JavaScript, with Codex's own output and context management. Appendices~\ref{app:prompt-texts} and~\ref{app:harness} document the prompts and implementation settings.

\section{Experiments}
\label{sec:experiments}

\begin{table*}[t]
\centering
\small
\colorlet{mainBase}{ovBlue!9}
\colorlet{mainOurs}{ovGreen!15}
\newcommand{\bandrule}{\specialrule{\lightrulewidth}{\aboverulesep}{\aboverulesep}}
\newcommand{\maingroup}[2]{\multicolumn{10}{@{}l@{}}{\rlap{\setlength{\fboxsep}{0pt}\colorbox{#1}{\makebox[\textwidth][c]{\rule[-3.3pt]{0pt}{11.2pt}\textit{#2}}}}}}
\setlength{\tabcolsep}{0pt}
\begin{tabular*}{\textwidth}{@{\extracolsep{\fill}}lccccccccc@{}}
\toprule
& \multicolumn{3}{c}{Amazon} & \multicolumn{3}{c}{MAG} & \multicolumn{3}{c}{Prime} \\
\cmidrule(lr){2-4}\cmidrule(lr){5-7}\cmidrule(l){8-10}
Method & \hone{} & \rtwenty{} & MRR & \hone{} & \rtwenty{} & MRR & \hone{} & \rtwenty{} & MRR \\
\bandrule
\maingroup{mainBase}{Trained or tuned on STaRK} \\
\midrule
AvaTaR~\citep{wu2024avatar} & 49.9 & 60.6 & 58.7 & 44.4 & 50.6 & 51.2 & 18.4 & 39.3 & 26.7 \\
PASemiQA~\citep{yang2025pasemiqa} & 45.9 & -- & 55.7 & 43.2 & -- & 50.2 & 29.7 & -- & 31.0 \\
mFAR (All+2)~\citep{li2024mfar} & 53.0 & 66.3 & 64.3 & 55.9 & 74.1 & 64.3 & 40.0 & 72.6 & 52.0 \\
MoR~\citep{lei2025mor} & 52.2 & 59.9 & 62.2 & 58.2 & 75.0 & 67.1 & 36.4 & 63.5 & 46.9 \\
GraphRAFT~\citep{clemedtson2025graphraft} & -- & -- & -- & 69.6 & 89.1 & 76.2 & 63.7 & 76.4 & 69.0 \\
GRASP~\citep{tao2026grasp} & 71.2 & 70.6 & 78.7 & 82.8 & 91.5 & 88.3 & 67.8 & \textbf{88.1} & 76.8 \\
\bandrule
\maingroup{mainBase}{Training-free} \\
\midrule
BM25~\citep{wu2024stark} & 44.9 & 53.8 & 55.3 & 25.9 & 45.7 & 34.9 & 12.8 & 31.3 & 19.8 \\
ColBERTv2~\citep{wu2024stark} & 46.1 & 53.4 & 55.5 & 31.2 & 43.9 & 38.4 & 11.8 & 25.0 & 17.4 \\
GritLM-7B~\citep{wu2024stark} & 42.1 & 56.5 & 53.5 & 37.9 & 46.4 & 47.3 & 15.6 & 39.1 & 24.1 \\
KAR~\citep{xia2024kar} & 54.2 & 57.2 & 61.3 & 50.5 & 60.3 & 57.5 & 30.4 & 50.8 & 39.2 \\
HybGRAG~\citep{lee2024hybgrag} & -- & -- & -- & 65.4 & 65.7 & 69.8 & 28.6 & 43.6 & 34.5 \\
AF-Retriever\smash{$^\dagger$}~\citep{boer2025afretriever} & 61.2 & 35.5 & 67.3 & 78.6 & 61.4 & 84.0 & 46.2 & 51.2 & 54.0 \\
ARK~\citep{polonuer2026ark} & 56.1 & 58.6 & 65.7 & 73.1 & 86.2 & 81.0 & 50.6 & 70.0 & 59.6 \\
\bandrule
\maingroup{mainBase}{KG reasoning methods adapted to STaRK by \citet{yang2025pasemiqa}, with GPT-4\smash{$^\ddagger$}} \\
\midrule
ToG~\citep{sun2024tog} & 16.7 & -- & 21.2 & 5.4 & -- & 6.1 & 13.2 & -- & 15.8 \\
RoG~\citep{luo2024rog} & 31.2 & -- & 36.0 & 32.9 & -- & 38.7 & 22.7 & -- & 25.2 \\
GoG~\citep{xu2024gog} & 21.0 & -- & 26.3 & 41.3 & -- & 47.8 & 18.9 & -- & 21.1 \\
\bandrule
\maingroup{mainOurs}{Ours: direct knowledge graph interaction (training-free)} \\
\midrule
\dkgi{} (Pi, GPT-6-Luna) & 57.8 & 58.1 & 64.7 & 88.0 & 88.4 & 89.8 & 63.8 & 78.9 & 70.4 \\
\dkgi{} (Codex, GPT-6-Luna) & 59.8 & 55.2 & 66.2 & 88.0 & 88.1 & 89.6 & 66.6 & 80.9 & 72.4 \\
\dkgi{} (Pi, GPT-6-Astra) & \textbf{79.4} & \textbf{71.1} & \textbf{84.0} & \textbf{91.8} & \textbf{93.2} & \textbf{93.2} & \textbf{74.5} & 86.1 & \textbf{79.1} \\
\bottomrule
\end{tabular*}
\caption{\hone{}, \rtwenty{} and MRR on the synthetic test sets of STaRK. Bold marks the best value in each column. Baseline numbers are published values (backbones in Appendix~\ref{app:baselines}), and a dash marks a value absent from the source paper. $^\dagger$\rtwenty{} from the complete-pipeline row of its ablation table. $^\ddagger$From the \hone{} and MRR tables of \citet{yang2025pasemiqa}. The \dkgi{} rows cover the full test sets (7,108 queries) and share the prompt and the time limit.}
\label{tab:main}
\vspace{-1em}
\end{table*}

\subsection{Setup}
\label{sec:setup}

\noindent \textbf{Data.} We use the full synthetic test sets of the STaRK benchmark~\citep{wu2024stark}, with 1,642, 2,665 and 2,801 queries for the Amazon, MAG and Prime domains, respectively (7,108 in total). We make all prompt and tool design decisions on validation queries and reserve test labels for evaluation. The ablations use the official 10\% test subset (164, 266 and 280 queries, 710 in total).

\smallskip
\noindent \textbf{Models and limits.} The default backbone is the lightweight GPT-6-Luna~\citep{openai2026gpt6luna}, and the stronger GPT-6-Astra~\citep{openai2026gpt6astra} is the second backbone, both with high reasoning effort. Each attempt has at most 1{,}800\,s of wall-clock time, and each Pi attempt at most 300 turns.

\smallskip
\noindent \textbf{Metrics.} Following STaRK, we report Hit@1 (\hone{}, also called Hits@1), Recall@20 (\rtwenty{}) and mean reciprocal rank (MRR). Because the agent is asked for 20 IDs, our MRR is a lower bound on the full-ranking MRR. A query whose attempts all fail counts as a miss in every metric (retry policy in Appendix~\ref{app:harness}). \emph{Overall} values are micro-averages over all queries of the three datasets, and differences are computed before rounding. Appendix~\ref{app:harness} reports the cost and resource use of each configuration.

\smallskip
\noindent \textbf{Baselines.} We report published results and, as in Table~\ref{tab:main}, separate the systems trained or tuned on STaRK from the training-free ones. The baselines are retrievers, LLM pipelines and agents, and all of them access the KG through predefined operations. The strongest trained system, GRASP~\citep{tao2026grasp}, combines a Cypher query on a graph database with a trained dense retriever and a fine-tuned reranker. Among the training-free systems, KAR~\citep{xia2024kar} and AF-Retriever~\citep{boer2025afretriever} are LLM pipelines. KAR expands the query from the documents of neighboring entities, and AF-Retriever translates it into graph constraints. The agent ARK~\citep{polonuer2026ark} calls predefined global search and one-hop neighborhood search tools. We list separately three KG reasoning methods (ToG, RoG and GoG) that \citet{yang2025pasemiqa} adapt to STaRK and run with GPT-4. Appendix~\ref{app:baselines} describes every system.

\subsection{Main Results}
\label{sec:results}

\begin{table*}[t]
\centering
\small
\setlength{\tabcolsep}{3.4pt}
\begin{tabular}{@{}lllcccccrcc@{}}
\toprule
& & & \multicolumn{5}{c}{\hone{}} & \multicolumn{2}{c}{Overall} \\
\cmidrule(lr){4-8}\cmidrule(lr){9-10}
Variant & Graph access & Python & Amazon & MAG & Prime & Overall & $\Delta$ & \rtwenty{} & MRR \\
\midrule
\vfull{} & node and edge files & \cmark{} & 59.1 & 91.0 & 69.6 & 75.2 & -- & 78.0 & 78.9 \\
\vshell{} & node and edge files & -- & 48.8 & 86.1 & 62.9 & 68.3 & $-6.9$ & 69.5 & 72.0 \\
\vtoolscode{} & predefined graph tools & \cmark{} & 60.4 & 90.2 & 67.1 & 74.2 & $-1.0$ & 79.4 & 78.9 \\
\vtools{} & predefined graph tools & -- & 59.8 & 87.6 & 66.4 & 72.8 & $-2.4$ & 79.5 & 77.6 \\
\midrule
\vone{} & node and edge files & once & 48.2 & 87.2 & 63.9 & 69.0 & $-6.2$ & 70.8 & 72.5 \\
\midrule
\vretr{} & \texttt{search}, \texttt{get\_node}, \texttt{get\_schema} & -- & 58.5 & 48.5 & 40.7 & 47.7 & $-27.5$ & 55.1 & 53.9 \\
\vnoedges{} & node files & \cmark{} & 49.4 & 45.5 & 40.0 & 44.2 & $-31.0$ & 52.6 & 50.6 \\
\vcodex{} & node and edge files & \cmark{} & 58.5 & 87.6 & 67.5 & 73.0 & $-2.3$ & 75.5 & 76.9 \\
\bottomrule
\end{tabular}
\caption{Ablation on the test subset (710 queries) with GPT-6-Luna in Pi, unless marked Codex. \emph{Graph access}: the graph files or graph tools that the agent can use. \emph{Python}: \cmark{} over several turns, ``once'' for a single program per query, and --~for no Python access. The first four rows form the $2\times2$ design, the fifth row limits Python to one program, and the last three rows are controls. $\Delta$: overall \hone{} minus that of \vfull{}.}
\label{tab:ablation}
\vspace{-1em}
\end{table*}

With GPT-6-Luna, \dkgi{} is the strongest training-free method on MAG and Prime, and with GPT-6-Astra it achieves the highest \hone{} and MRR on every dataset (Table~\ref{tab:main}). On MAG and Prime, \dkgi{} with GPT-6-Luna in Pi outperforms every training-free baseline in all three metrics and exceeds the best of them in \hone{} by 9.4 points on MAG (AF-Retriever) and by 13.2 points on Prime (ARK). On every dataset, it also outperforms the three KG reasoning methods that \citet{yang2025pasemiqa} adapt to STaRK, whose \hone{} is at most 41.3.

As a training-free method, \dkgi{} also outperforms the strongest trained system, GRASP, in \hone{} and MRR, on MAG with GPT-6-Luna and on all three datasets with GPT-6-Astra. With GPT-6-Luna in Pi, its margins on MAG are 5.2 \hone{} and 1.5 MRR points. With GPT-6-Astra in Pi, its \hone{} margins over GRASP are 8.2, 9.0 and 6.7 points on Amazon, MAG and Prime, respectively. With the same prompt and harness, overall \hone{} rises from 71.5 with GPT-6-Luna to 82.1 with GPT-6-Astra.

The results carry over to a second agent harness. In Codex, with the same backbone (GPT-6-Luna), prompt and time limit, \dkgi{} reaches an overall \hone{} of 73.1 on the full test sets, compared with 71.5 in Pi (Table~\ref{tab:main}). Appendix~\ref{app:ablation} gives the comparison on the test subset.

\subsection{Ablation Study}
\label{sec:ablation}

Section~\ref{sec:results} compares \dkgi{} with published systems that differ from it in model, training and graph access. To identify which components of \dkgi{} account for its accuracy, we fix the model (GPT-6-Luna), the harness (Pi), the budgets and the prompt apart from its tool section (Appendices~\ref{app:prompt-texts} and~\ref{app:harness}), and vary which actions $a_t$ (\S\ref{sec:method}) the agent can take. All variants run on the test subset (Table~\ref{tab:ablation}).

The $2\times2$ design crosses graph access with Python. \emph{With predefined graph tools.} The agent has four graph tools based on ARK~\citep{polonuer2026ark}. \texttt{search} ranks nodes by BM25 over their names and summaries, \texttt{get\_node} returns the record of an ID, \texttt{get\_schema} returns the schema, and \texttt{neighbors} returns at most 20 one-hop neighbors of a node, with optional relation, type and keyword filters. In \vtools{}, each graph access is a single call to one of these tools, as in ARK, Graph-CoT and ToG~\citep{polonuer2026ark,jin2024graphcot,sun2024tog}. In \vtoolscode{}, the agent writes and revises Python programs that reach the graph through client calls to the four tools. AvaTaR and KG-Agent also compose predefined tool calls in code~\citep{wu2024avatar,jiang2024kgagent}. \emph{With the graph files.} \vfull{} has the graph files, a shell and Python. In \vshell{}, the agent combines Pi's \texttt{read} tool and shell commands such as \texttt{rg}, \texttt{sed} and \texttt{sort} with bash pipelines, loops and arithmetic, while \texttt{python3}, \texttt{awk}, \texttt{perl} and \texttt{node} are blocked. \vone{} restricts \vfull{} to the shell commands of \vshell{} and a single Python run per query, so the agent writes all of its Python operations before it sees the output of any of them. \emph{Control variants.} \vretr{} removes \texttt{neighbors} from \vtools{}, and \vnoedges{} removes the edge files from \vfull{}. In the closed-book probes (Appendix~\ref{app:probes}), the model answers once from its own knowledge, with STaRK node IDs (the \emph{ID probe}) or, on MAG and Prime, with entity names mapped to IDs offline (the \emph{name probe}).

\begin{figure*}[t]
\centering
\includegraphics[width=0.98\textwidth]{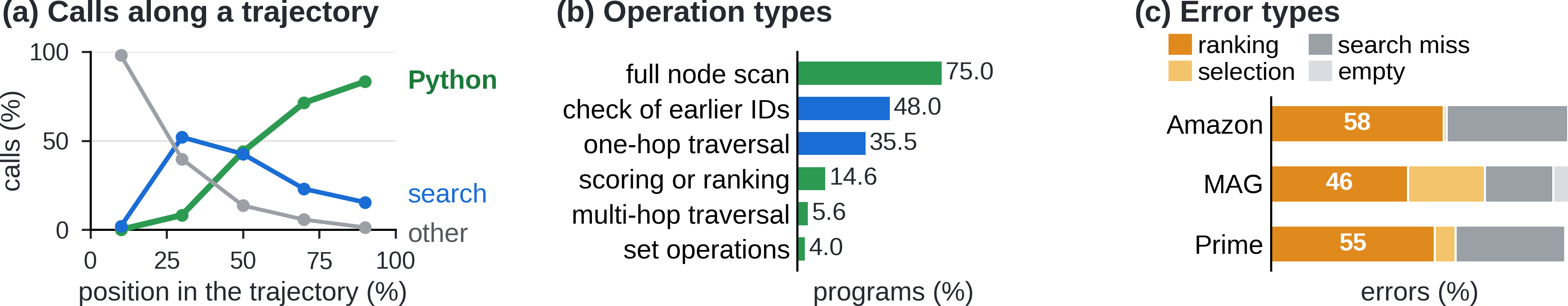}
\caption{Behavior of \vfull{} (Pi, GPT-6-Luna) on the full test sets. (a)~Python commands, \texttt{rg} searches and other calls (mostly \texttt{read}) by position in the trajectory. (b)~Programs by operation type (types overlap). The blue types have counterparts among predefined graph tools. (c)~Error types of incorrectly answered queries (Appendix~\ref{app:errors}).}
\label{fig:coevolution}
\vspace{-1em}
\end{figure*}

\smallskip
\noindent \textbf{With Python, the agent outperforms predefined graph tools in overall \hone{}.}
With the shell but no Python (\vshell{}), the agent scores 68.3 overall \hone{}, below \vtools{} (72.8), and it is also lower on every dataset and in overall \rtwenty{} and MRR. It prints entire records (an Amazon product record includes all of its reviews) and follows relations by searching the edge files with \texttt{rg}. With Python, \vfull{} reaches 75.2, 2.4 points above \vtools{}, and its MRR is also higher (78.9 vs.\ 77.6). The lead comes from MAG and Prime.

\smallskip
\noindent \textbf{Python improves the agent far more than it improves predefined graph tools.}
On the graph files, adding Python (\vfull{} vs.\ \vshell{}) improves overall \hone{} by 6.9 points, \rtwenty{} by 8.5 and MRR by 6.9. With the graph tools, adding Python (\vtoolscode{} vs.\ \vtools{}) improves overall \hone{} and MRR by 1.4 points each. The tools \texttt{search} and \texttt{neighbors} already return compact entries with node IDs, names and types. On the graph files, the agent's programs produce such entries themselves, by extracting a few fields from each node record and following relations over the edge files.

\smallskip
\noindent \textbf{The agent gains most from Python when it writes and executes programs over several turns.}
\vone{} reaches 69.0 overall \hone{}, 0.7 points above \vshell{} and 6.2 below \vfull{}. A single program recovers about a tenth of the 6.9-point gain from Python on the graph files, although the agent runs its program in 81\% of the queries. On the queries in which both variants run Python, this program is longer than the first program of \vfull{} (median 15 vs.\ 8 lines) and shorter than all programs of \vfull{} on the query together (median 41 lines). These results support the value of repeated, adaptive programming.

\smallskip
\noindent \textbf{\vfull{} achieves higher overall \hone{} than predefined graph tools composed in code.}
\vfull{} is 1.0 point above \vtoolscode{} in overall \hone{} (75.2 vs.\ 74.2), and the two are equal in MRR (78.9). On MAG and Prime together, its \hone{} is 80.0, compared with 76.7 for \vtools{} and 78.4 for \vtoolscode{}. Unlike both tool variants, it needs no predefined graph API and works on the exported graph files.

\smallskip
\noindent \textbf{The agent relies on the relations in the graph.}
Removing access to the relations lowers accuracy sharply, mostly on MAG and Prime. \vnoedges{} reaches 44.2 overall \hone{}, 31.0 points below \vfull{}, and \vretr{} reaches 47.7, 25.1 points below \vtools{}. The ID probe shows little recall of benchmark IDs. On the full test sets, the ID probe answers one Prime query correctly, and the name probe answers 2.3\% of MAG and 20.9\% of Prime queries correctly. On the 223 test-subset Prime queries that the name probe misses, the lead of \vfull{} over \vtools{} is 4.0 points (63.2 vs.\ 59.2), compared with 3.2 points on all Prime queries of the test subset.

\subsection{How Tool Creation and Tool Use Co-Evolve}
\label{sec:analysis}

Section~\ref{sec:ablation} shows that the agent gains most from Python when it writes and executes programs over several turns. We trace this process in the 7,108 trajectories of \vfull{} with GPT-6-Luna on the full test sets (Figure~\ref{fig:coevolution}).

\smallskip
\noindent \textbf{The agent searches first and then writes programs.} A trajectory starts with file reads and \texttt{rg} searches, and Python makes up 8.2\% of the calls between 20\% and 40\% of a trajectory and 83.3\% of the calls in its last 20\% (Figure~\ref{fig:coevolution}a). The agent writes 4.0 programs per query on average. Of all programs, 75.0\% contain a full node scan and 35.5\% a one-hop traversal. In 42.3\% of the queries, a program contains scoring or ranking, a multi-hop traversal or a set operation (Figure~\ref{fig:coevolution}b). The two operation types in blue have counterparts among the predefined graph tools, as \texttt{neighbors} performs a one-hop traversal and \texttt{get\_node} looks up a node by its ID.

\smallskip
\noindent \textbf{Programs reuse IDs from earlier outputs and code from earlier programs.} Of all programs, 48.4\% contain a node ID shown in the output of an earlier turn, and almost all of these programs check the IDs (48.0\% of all programs, Figure~\ref{fig:coevolution}b). In the request of Figure~\ref{fig:overview}, two \texttt{rg} searches return the IDs of the university and of the field \emph{Data quality}, and the agent's programs take these IDs as input to collect the 236 papers of the university's 66 authors and intersect them with the 86 papers of the field. Programs also reuse the code of earlier programs. Of all programs, 36.5\% revise an earlier one under our similarity criterion (Appendix~\ref{app:behaviour}). After a program fails, the next program revises it in 88.6\% of cases, and after a program runs successfully, the next program revises it in 49.7\% of cases. In this way, the agent's tools evolve from one another within a task. \vone{}, which allows a single program per query, removes this process and recovers a tenth of the gain from Python (\S\ref{sec:ablation}).

\smallskip
\noindent \textbf{The agent's lead over predefined graph tools is largest on queries where both tool-based agents make many calls.} On the 100 test-subset queries for which both \vtools{} and \vtoolscode{} call \texttt{neighbors} at least ten times, \vfull{} leads them by 9.0 and 6.0 \hone{} points, compared with 1.3 and 0.2 points on the other 610 queries. These 100 queries all come from MAG and Prime, and 85.0\% of them have a program with scoring or ranking, a multi-hop traversal or a set operation, compared with 37.5\% of the other queries. The request of Figure~\ref{fig:overview} is one of these 100 queries. The \vtools{} agent misses its gold paper, which \vfull{} ranks first (Appendix~\ref{app:cases} compares the two agents on two more requests).

\smallskip
\noindent \textbf{With a stronger LLM, the agent reuses earlier outputs more often and revises more of its failed programs.} With GPT-6-Astra, the agent turns to programs earlier (Python makes up 31.4\% of the calls between 20\% and 40\% of a trajectory, compared with 8.2\%), 59.9\% of its programs contain an ID from an earlier output (compared with 48.4\%), and the next program revises a failed one in 92.2\% of cases (compared with 88.6\%). With GPT-6-Luna in Codex, the last two shares are 50.3\% and 85.1\%, close to those in Pi (Appendix~\ref{app:behaviour}).

\smallskip
\noindent \textbf{Ranking is the largest source of remaining errors.} Among the queries that \vfull{} answers incorrectly, a gold node is returned but ranked below the first position in 58\%, 46\% and 55\% on Amazon, MAG and Prime, and search misses account for 41\%, 23\% and 37\% (Figure~\ref{fig:coevolution}c).

\section{Related Work}
\label{sec:related}

\noindent \textbf{Direct tool use.}
Retrieve-then-reason methods~\citep{jiang2023unikgqa,luo2024rog,kim2023kggpt,he2024gretriever,mavromatis2025gnnrag,li2025subgraphrag} retrieve reasoning paths or subgraphs from the KG and reason over them. In KG agents such as ToG~\citep{sun2024tog,ma2025tog2}, the LLM steers a predefined beam search over entities and relations, and StructGPT~\citep{jiang2023structgpt}, Graph-CoT~\citep{jin2024graphcot}, GoG~\citep{xu2024gog} and Plan-on-Graph~\citep{chen2024pog} give it other sets of predefined graph functions. GRA~\citep{dragic2026gra} explores hybrid KGs through seven predefined Unix-style tools such as \texttt{ls}, \texttt{cat} and \texttt{grep}.

\smallskip
\noindent \textbf{Code over a graph interface.}
Semantic parsing maps a question to SPARQL, an S-expression or a program executed on the KG~\citep{liang2017nsm,cao2022kqapro,li2023kbbinder,gu2023pangu,luo2024chatkbqa}, and \citet{xiong2024interactivekbqa} build these logical forms interactively through knowledge-base APIs. KG-Agent~\citep{jiang2024kgagent} fine-tunes a small model to generate programs that call predefined tools. Code-on-Graph~\citep{ding2026codeongraph} represents the schema as Python classes and generates code that processes graph objects provided by a subgraph retriever. On STaRK, AvaTaR~\citep{wu2024avatar} optimizes the prompt of an agent that composes predefined functions in Python.

\smallskip
\noindent \textbf{Agents that create their own tools.}
Outside KGs, CodeAct agents~\citep{wang2024codeact} act through executable Python, and other methods create tools as code~\citep{cai2024latm,yuan2024craft,wang2024trove,wolflein2025toolmaker}. CREATOR~\citep{qian2023creator}, Self-Debug~\citep{chen2024selfdebug} and PyVision~\citep{zhao2025pyvision} refine code using execution feedback. Live-SWE-agent~\citep{xia2025liveswe} creates tools within a minimal bash scaffold to solve software issues. SMITH~\citep{tam2026smith} jointly trains tool creation and use with reinforcement learning and reuses tools across tasks, whereas \dkgi{} is training-free and creates and revises tools within a task. Graph-as-Code~\citep{finkelshtein2026graphascode} iteratively writes and executes code over a table of node features, neighbor lists and labels for node classification. For entity search over typed, multi-relational KGs, \dkgi{} is to our knowledge the first harness that lets agents implement graph operations over complete node and edge files.

\section{Conclusion}
\label{sec:conclusion}
We presented \dkgi{}, a harness where LLM agents write and execute their own graph access code over the complete node and edge files. With GPT-6-Astra, the training-free \dkgi{} achieves the highest \hone{} and MRR on every STaRK dataset. In controlled ablations with GPT-6-Luna, the agent achieves higher overall \hone{} than predefined graph tools, whether called directly or composed in code, and it gains most when it writes and revises programs over several turns. Its trajectories show how tool creation and tool use co-evolve within a task.

\section*{Limitations}
\noindent \textbf{Scope.} Our evaluation covers the three text-rich KGs of STaRK, which have at most 10 entity types, 18 relation types, 1.9M entities and 39.8M edges~\citep{wu2024stark}. Other KG benchmarks such as WebQSP, CWQ or GrailQA~\citep{yih2016webqsp,talmor2018cwq,gu2021grailqa}, graphs with Freebase- or Wikidata-scale schemas, aggregation questions and the scaling of \dkgi{} with graph size are left to future work. Where a KG is too large to scan or a question needs exact aggregation, purpose-built indexes, graph databases and query languages may be more suitable, and an agent could write its graph operations on top of them. A comparison with an agent in the same harness that queries a graph database through a query language such as Cypher or SPARQL~\citep{feng2024cypherbench} is also left to future work. GraphRAFT, which fine-tunes a text-to-Cypher model, enters the comparison through its published numbers.

\smallskip
\noindent \textbf{Models, baselines and data.} Our main results use the proprietary models GPT-6-Luna and GPT-6-Astra, which we accessed through an API in September and October 2026. Exact reproduction depends on the continued availability of these models. Open-weight models remain to be tested. We release the code under the MIT License and the exported graphs, prompts and all trajectories under CC BY 4.0. Baseline numbers are taken from the literature and were obtained with different backbones, budgets and evaluation code. The ablations in \S\ref{sec:ablation} hold these factors fixed. Our evaluation uses STaRK's synthetic queries, which make up most of the benchmark.

\smallskip
\noindent \textbf{Protocol and deployment.} The agent's programs are written inline and discarded after each question. Because the agent runs arbitrary shell commands, any deployment requires a sandbox such as ours (Appendix~\ref{app:harness}).

\section*{Ethical Considerations}
We use the public STaRK benchmark~\citep{wu2024stark}, whose data are released under CC BY 4.0. The code of STaRK, DCI-Agent-Lite~\citep{li2026beyond}, the Pi coding agent and ARK is released under the MIT License, and Codex CLI under the Apache License 2.0. Our research use of these artifacts is consistent with their licenses, and our derived data keep STaRK's CC BY 4.0 license. The code, exported graphs, prompts and trajectories that we release are intended for research on agents that search knowledge graphs. STaRK's queries are in English, as are almost all node records. The Amazon data contain user reviews, and the MAG data contain author names and affiliations. These public data are part of the benchmark. We use them as provided by STaRK, and all personal data in this work come from the benchmark. Because the agent can execute arbitrary shell commands, we ran it in a sandbox that keeps the graph files read-only and blocks access to the network and the labels (Appendix~\ref{app:harness}). We report point estimates from one evaluation per configuration and query set, with failed attempts retried as described in Appendix~\ref{app:harness}. Appendix~\ref{app:harness} gives the computational cost. The Amazon and Prime text-constraint tags in Appendix~\ref{app:behaviour} were assigned by an LLM annotator, GPT-6-Luna.
We used AI assistants for coding and for writing assistance.
\vfill

\bibliography{custom}

\begin{thebibliography}{67}
\providecommand{\natexlab}[1]{#1}

\bibitem[{Boer et~al.(2026)Boer, Roth, and Kramer}]{boer2025afretriever}
Derian Boer, Stephen Roth, and Stefan Kramer. 2026.
\newblock \href {https://arxiv.org/abs/2505.09246v4} {{Autofocus Retrieval: An Effective Pipeline for Multi-Hop Question Answering With Semi-Structured Knowledge}}.
\newblock \emph{Transactions on Machine Learning Research}.

\bibitem[{Cai et~al.(2024)Cai, Wang, Ma, Chen, and Zhou}]{cai2024latm}
Tianle Cai, Xuezhi Wang, Tengyu Ma, Xinyun Chen, and Denny Zhou. 2024.
\newblock \href {https://arxiv.org/abs/2305.17126v2} {Large language models as tool makers}.
\newblock In \emph{International Conference on Learning Representations}, pages 54067--54089.

\bibitem[{Cao et~al.(2022)Cao, Shi, Pan, Nie, Xiang, Hou, Li, He, and Zhang}]{cao2022kqapro}
Shulin Cao, Jiaxin Shi, Liangming Pan, Lunyiu Nie, Yutong Xiang, Lei Hou, Juanzi Li, Bin He, and Hanwang Zhang. 2022.
\newblock \href {https://doi.org/10.18653/v1/2022.acl-long.422} {{{KQA} Pro: A Dataset with Explicit Compositional Programs for Complex Question Answering over Knowledge Base}}.
\newblock In \emph{Proceedings of the 60th Annual Meeting of the Association for Computational Linguistics (Volume 1: Long Papers)}, pages 6101--6119.

\bibitem[{Chandak et~al.(2023)Chandak, Huang, and Zitnik}]{chandak2023primekg}
Payal Chandak, Kexin Huang, and Marinka Zitnik. 2023.
\newblock \href {https://doi.org/10.1038/s41597-023-01960-3} {{Building a knowledge graph to enable precision medicine}}.
\newblock \emph{Scientific Data}, 10(1):67.

\bibitem[{Chen et~al.(2024{\natexlab{a}})Chen, Tong, Jin, Sun, Ye, and Xiong}]{chen2024pog}
Liyi Chen, Panrong Tong, Zhongming Jin, Ying Sun, Jieping Ye, and Hui Xiong. 2024{\natexlab{a}}.
\newblock \href {https://arxiv.org/abs/2410.23875} {Plan-on-graph: Self-correcting adaptive planning of large language model on knowledge graphs}.
\newblock In \emph{Advances in Neural Information Processing Systems}.

\bibitem[{Chen et~al.(2024{\natexlab{b}})Chen, Lin, Sch{\"a}rli, and Zhou}]{chen2024selfdebug}
Xinyun Chen, Maxwell Lin, Nathanael Sch{\"a}rli, and Denny Zhou. 2024{\natexlab{b}}.
\newblock \href {https://arxiv.org/abs/2304.05128v2} {Teaching large language models to self-debug}.
\newblock In \emph{International Conference on Learning Representations}, pages 8746--8825.

\bibitem[{Chen et~al.(2026)Chen, Ma, Zhuang, Nie, Zou, Sharifymoghaddam, Liu, Green, Patel, Meng, Su, Li, Hong, Shi, Liu, Oyarhoseini, Thakur, Zhang, Gao, Chen, and Lin}]{chen2025browsecompplus}
Zijian Chen, Xueguang Ma, Shengyao Zhuang, Ping Nie, Kai Zou, Sahel Sharifymoghaddam, Andrew Liu, Joshua Green, Kshama Patel, Ruoxi Meng, Mingyi Su, Yanxi Li, Haoran Hong, Xinyu Shi, Xuye Liu, Hosna Oyarhoseini, Nandan Thakur, Crystina Zhang, Luyu Gao, and 2 others. 2026.
\newblock \href {https://doi.org/10.18653/v1/2026.acl-long.1023} {{BrowseComp-Plus: A Fair and Disentangled Evaluation Benchmark for Deep Search Agents}}.
\newblock In \emph{Proceedings of the 64th Annual Meeting of the Association for Computational Linguistics (Volume 1: Long Papers)}, pages 22349--22370.

\bibitem[{Clemedtson and Shi(2025)}]{clemedtson2025graphraft}
Alfred Clemedtson and Borun Shi. 2025.
\newblock \href {https://arxiv.org/abs/2504.05478v2} {{GraphRAFT: Retrieval Augmented Fine-Tuning for Knowledge Graphs on Graph Databases}}.
\newblock \emph{Preprint}, arXiv:2504.05478v2.

\bibitem[{Ding et~al.(2026)Ding, Li, Bai, Chen, Su, Wang, Jin, Zhang, Guo, and Cheng}]{ding2026codeongraph}
Weiwei Ding, Zixuan Li, Long Bai, Zhuo Chen, Kun Su, Fei Wang, Xiaolong Jin, Jin Zhang, Jiafeng Guo, and Xueqi Cheng. 2026.
\newblock \href {https://arxiv.org/abs/2606.03705} {{Code-on-Graph: Iterative Programmatic Reasoning via Large Language Models on Knowledge Graphs}}.
\newblock \emph{arXiv preprint arXiv:2606.03705}.

\bibitem[{Dragic et~al.(2026)Dragic, Ifrah, and Rio}]{dragic2026gra}
Marius Dragic, Ruben Ifrah, and Alexandre Rio. 2026.
\newblock \href {https://arxiv.org/abs/2608.15834} {Schema-agnostic graph reasoning agent for hybrid knowledge graphs}.
\newblock \emph{arXiv preprint arXiv:2608.15834}.

\bibitem[{Feng et~al.(2025)Feng, Papicchio, and Rahman}]{feng2024cypherbench}
Yanlin Feng, Simone Papicchio, and Sajjadur Rahman. 2025.
\newblock \href {https://doi.org/10.18653/v1/2025.acl-long.438} {{CypherBench}: Towards precise retrieval over full-scale modern knowledge graphs in the {LLM} era}.
\newblock In \emph{Proceedings of the 63rd Annual Meeting of the Association for Computational Linguistics (Volume 1: Long Papers)}, pages 8934--8958, Vienna, Austria. Association for Computational Linguistics.

\bibitem[{Finkelshtein et~al.(2026)Finkelshtein, Cucerzan, Jauhar, and White}]{finkelshtein2026graphascode}
Ben Finkelshtein, Silviu Cucerzan, Sujay~Kumar Jauhar, and Ryen White. 2026.
\newblock \href {https://arxiv.org/abs/2509.18487} {Actions speak louder than prompts: A large-scale study of {LLMs} for graph inference}.
\newblock In \emph{International Conference on Learning Representations}.

\bibitem[{Gu et~al.(2023)Gu, Deng, and Su}]{gu2023pangu}
Yu~Gu, Xiang Deng, and Yu~Su. 2023.
\newblock \href {https://doi.org/10.18653/v1/2023.acl-long.270} {{Don{'}t Generate, Discriminate: A Proposal for Grounding Language Models to Real-World Environments}}.
\newblock In \emph{Proceedings of the 61st Annual Meeting of the Association for Computational Linguistics (Volume 1: Long Papers)}, pages 4928--4949.

\bibitem[{Gu et~al.(2021)Gu, Kase, Vanni, Sadler, Liang, Yan, and Su}]{gu2021grailqa}
Yu~Gu, Sue Kase, Michelle Vanni, Brian Sadler, Percy Liang, Xifeng Yan, and Yu~Su. 2021.
\newblock \href {https://arxiv.org/abs/2011.07743} {Beyond {I.I.D.}: Three levels of generalization for question answering on knowledge bases}.
\newblock In \emph{Proceedings of the Web Conference 2021}.

\bibitem[{He and McAuley(2016)}]{he2016ups}
Ruining He and Julian McAuley. 2016.
\newblock \href {https://doi.org/10.1145/2872427.2883037} {{Ups and Downs: Modeling the Visual Evolution of Fashion Trends with One-Class Collaborative Filtering}}.
\newblock In \emph{Proceedings of the 25th International Conference on World Wide Web}, pages 507--517.

\bibitem[{He et~al.(2024)He, Tian, Sun, Chawla, Laurent, LeCun, Bresson, and Hooi}]{he2024gretriever}
Xiaoxin He, Yijun Tian, Yifei Sun, Nitesh~V. Chawla, Thomas Laurent, Yann LeCun, Xavier Bresson, and Bryan Hooi. 2024.
\newblock \href {https://arxiv.org/abs/2402.07630} {{G-Retriever}: Retrieval-augmented generation for textual graph understanding and question answering}.
\newblock In \emph{Advances in Neural Information Processing Systems}.

\bibitem[{Hu et~al.(2020)Hu, Fey, Zitnik, Dong, Ren, Liu, Catasta, and Leskovec}]{hu2020ogb}
Weihua Hu, Matthias Fey, Marinka Zitnik, Yuxiao Dong, Hongyu Ren, Bowen Liu, Michele Catasta, and Jure Leskovec. 2020.
\newblock \href {https://proceedings.neurips.cc/paper/2020/hash/fb60d411a5c5b72b2e7d3527cfc84fd0-Abstract.html} {{Open Graph Benchmark: Datasets for Machine Learning on Graphs}}.
\newblock In \emph{Advances in Neural Information Processing Systems}, volume~33, pages 22118--22133.

\bibitem[{Jiang et~al.(2023{\natexlab{a}})Jiang, Zhou, Dong, Ye, Zhao, and Wen}]{jiang2023structgpt}
Jinhao Jiang, Kun Zhou, Zican Dong, Keming Ye, Wayne~Xin Zhao, and Ji-Rong Wen. 2023{\natexlab{a}}.
\newblock \href {https://doi.org/10.18653/v1/2023.emnlp-main.574} {{{S}truct{GPT}: A General Framework for Large Language Model to Reason over Structured Data}}.
\newblock In \emph{Proceedings of the 2023 Conference on Empirical Methods in Natural Language Processing}, pages 9237--9251.

\bibitem[{Jiang et~al.(2025)Jiang, Zhou, Zhao, Song, Zhu, Zhu, and Wen}]{jiang2024kgagent}
Jinhao Jiang, Kun Zhou, Wayne~Xin Zhao, Yang Song, Chen Zhu, Hengshu Zhu, and Ji-Rong Wen. 2025.
\newblock \href {https://doi.org/10.18653/v1/2025.acl-long.468} {{KG-Agent: An Efficient Autonomous Agent Framework for Complex Reasoning over Knowledge Graph}}.
\newblock In \emph{Proceedings of the 63rd Annual Meeting of the Association for Computational Linguistics (Volume 1: Long Papers)}, pages 9505--9523.

\bibitem[{Jiang et~al.(2023{\natexlab{b}})Jiang, Zhou, Zhao, and Wen}]{jiang2023unikgqa}
Jinhao Jiang, Kun Zhou, Wayne~Xin Zhao, and Ji-Rong Wen. 2023{\natexlab{b}}.
\newblock \href {https://arxiv.org/abs/2212.00959} {{UniKGQA}: Unified retrieval and reasoning for solving multi-hop question answering over knowledge graph}.
\newblock In \emph{International Conference on Learning Representations}.

\bibitem[{Jin et~al.(2024)Jin, Xie, Zhang, Roy, Zhang, Li, Li, Tang, Wang, Meng, and Han}]{jin2024graphcot}
Bowen Jin, Chulin Xie, Jiawei Zhang, Kashob~Kumar Roy, Yu~Zhang, Zheng Li, Ruirui Li, Xianfeng Tang, Suhang Wang, Yu~Meng, and Jiawei Han. 2024.
\newblock \href {https://doi.org/10.18653/v1/2024.findings-acl.11} {{Graph Chain-of-Thought: Augmenting Large Language Models by Reasoning on Graphs}}.
\newblock In \emph{Findings of the Association for Computational Linguistics: ACL 2024}, pages 163--184.

\bibitem[{Kim et~al.(2023)Kim, Kwon, Jo, and Choi}]{kim2023kggpt}
Jiho Kim, Yeonsu Kwon, Yohan Jo, and Edward Choi. 2023.
\newblock \href {https://doi.org/10.18653/v1/2023.findings-emnlp.631} {{{KG}-{GPT}: A General Framework for Reasoning on Knowledge Graphs Using Large Language Models}}.
\newblock In \emph{Findings of the Association for Computational Linguistics: EMNLP 2023}, pages 9410--9421.

\bibitem[{Lee et~al.(2025)Lee, Zhu, Mavromatis, Han, Adeshina, Ioannidis, Rangwala, and Faloutsos}]{lee2024hybgrag}
Meng-Chieh Lee, Qi~Zhu, Costas Mavromatis, Zhen Han, Soji Adeshina, Vassilis~N. Ioannidis, Huzefa Rangwala, and Christos Faloutsos. 2025.
\newblock \href {https://doi.org/10.18653/v1/2025.acl-long.43} {{HybGRAG: Hybrid Retrieval-Augmented Generation on Textual and Relational Knowledge Bases}}.
\newblock In \emph{Proceedings of the 63rd Annual Meeting of the Association for Computational Linguistics (Volume 1: Long Papers)}, pages 879--893.

\bibitem[{Lei et~al.(2025)Lei, Han, Rossi, Dernoncourt, Lipka, Halappanavar, Tang, and Wang}]{lei2025mor}
Yongjia Lei, Haoyu Han, Ryan~A. Rossi, Franck Dernoncourt, Nedim Lipka, Mahantesh~M Halappanavar, Jiliang Tang, and Yu~Wang. 2025.
\newblock \href {https://doi.org/10.18653/v1/2025.findings-acl.941} {{Mixture of Structural-and-Textual Retrieval over Text-rich Graph Knowledge Bases}}.
\newblock In \emph{Findings of the Association for Computational Linguistics: ACL 2025}, pages 18306--18321.

\bibitem[{Li et~al.(2025{\natexlab{a}})Li, Chen, Van~Durme, and Xia}]{li2024mfar}
Millicent Li, Tongfei Chen, Benjamin Van~Durme, and Patrick Xia. 2025{\natexlab{a}}.
\newblock \href {https://arxiv.org/abs/2410.20056v2} {{Multi-Field Adaptive Retrieval}}.
\newblock In \emph{International Conference on Learning Representations}.

\bibitem[{Li et~al.(2025{\natexlab{b}})Li, Miao, and Li}]{li2025subgraphrag}
Mufei Li, Siqi Miao, and Pan Li. 2025{\natexlab{b}}.
\newblock \href {https://arxiv.org/abs/2410.20724} {Simple is effective: The roles of graphs and large language models in knowledge-graph-based retrieval-augmented generation}.
\newblock In \emph{International Conference on Learning Representations}.

\bibitem[{Li et~al.(2023)Li, Ma, Zhuang, Gu, Su, and Chen}]{li2023kbbinder}
Tianle Li, Xueguang Ma, Alex Zhuang, Yu~Gu, Yu~Su, and Wenhu Chen. 2023.
\newblock \href {https://doi.org/10.18653/v1/2023.acl-long.385} {{Few-shot In-context Learning on Knowledge Base Question Answering}}.
\newblock In \emph{Proceedings of the 61st Annual Meeting of the Association for Computational Linguistics (Volume 1: Long Papers)}, pages 6966--6980.

\bibitem[{Li et~al.(2026)Li, Zhang, Wei, Lu, Nie, Lu, Bai, Feng, Zhu, Zhong, Zhang, Xie, Choi, Zou, Han, Chen, Lin, Jiang, and Zhang}]{li2026beyond}
Zhuofeng Li, Haoxiang Zhang, Cong Wei, Pan Lu, Ping Nie, Yi~Lu, Yuyang Bai, Shangbin Feng, Hangxiao Zhu, Ming Zhong, Yuyu Zhang, Jianwen Xie, Yejin Choi, James Zou, Jiawei Han, Wenhu Chen, Jimmy Lin, Dongfu Jiang, and Yu~Zhang. 2026.
\newblock \href {https://arxiv.org/abs/2605.05242v1} {Beyond semantic similarity: Rethinking retrieval for agentic search via direct corpus interaction}.
\newblock In \emph{Advances in Neural Information Processing Systems}.

\bibitem[{Liang et~al.(2017)Liang, Berant, Le, Forbus, and Lao}]{liang2017nsm}
Chen Liang, Jonathan Berant, Quoc Le, Kenneth~D. Forbus, and Ni~Lao. 2017.
\newblock \href {https://doi.org/10.18653/v1/P17-1003} {{Neural Symbolic Machines: Learning Semantic Parsers on {F}reebase with Weak Supervision}}.
\newblock In \emph{Proceedings of the 55th Annual Meeting of the Association for Computational Linguistics (Volume 1: Long Papers)}, pages 23--33.

\bibitem[{Lu et~al.(2026)Lu, Li, Nie, Zhang, Zhang, Zou, Chen, Lin, Jiang, and Zhang}]{lu2026drdci}
Yi~Lu, Zhuofeng Li, Ping Nie, Haoxiang Zhang, Yuyu Zhang, Kai Zou, Wenhu Chen, Jimmy Lin, Dongfu Jiang, and Yu~Zhang. 2026.
\newblock \href {https://arxiv.org/abs/2606.14885} {{Dr-DCI}: Scaling direct corpus interaction via dynamic workspace expansion}.
\newblock \emph{arXiv preprint arXiv:2606.14885}.

\bibitem[{Luo et~al.(2024{\natexlab{a}})Luo, E, Tang, Peng, Guo, Zhang, Ma, Dong, Song, Lin, Zhu, and Luu}]{luo2024chatkbqa}
Haoran Luo, Haihong E, Zichen Tang, Shiyao Peng, Yikai Guo, Wentai Zhang, Chenghao Ma, Guanting Dong, Meina Song, Wei Lin, Yifan Zhu, and Anh~Tuan Luu. 2024{\natexlab{a}}.
\newblock \href {https://doi.org/10.18653/v1/2024.findings-acl.122} {{{C}hat{KBQA}: A Generate-then-Retrieve Framework for Knowledge Base Question Answering with Fine-tuned Large Language Models}}.
\newblock In \emph{Findings of the Association for Computational Linguistics: ACL 2024}, pages 2039--2056.

\bibitem[{Luo et~al.(2024{\natexlab{b}})Luo, Li, Haffari, and Pan}]{luo2024rog}
Linhao Luo, Yuan-Fang Li, Gholamreza Haffari, and Shirui Pan. 2024{\natexlab{b}}.
\newblock \href {https://arxiv.org/abs/2310.01061} {Reasoning on graphs: Faithful and interpretable large language model reasoning}.
\newblock In \emph{International Conference on Learning Representations}.

\bibitem[{Ma et~al.(2025)Ma, Xu, Jiang, Li, Qu, Yang, Mao, and Guo}]{ma2025tog2}
Shengjie Ma, Chengjin Xu, Xuhui Jiang, Muzhi Li, Huaren Qu, Cehao Yang, Jiaxin Mao, and Jian Guo. 2025.
\newblock \href {https://arxiv.org/abs/2407.10805} {Think-on-graph 2.0: Deep and faithful large language model reasoning with knowledge-guided retrieval augmented generation}.
\newblock In \emph{International Conference on Learning Representations}.

\bibitem[{Mavromatis and Karypis(2025)}]{mavromatis2025gnnrag}
Costas Mavromatis and George Karypis. 2025.
\newblock \href {https://doi.org/10.18653/v1/2025.findings-acl.856} {{{GNN}-{RAG}: Graph Neural Retrieval for Efficient Large Language Model Reasoning on Knowledge Graphs}}.
\newblock In \emph{Findings of the Association for Computational Linguistics: ACL 2025}, pages 16682--16699.

\bibitem[{McAuley et~al.(2015)McAuley, Targett, Shi, and van~den Hengel}]{mcauley2015image}
Julian McAuley, Christopher Targett, Qinfeng Shi, and Anton van~den Hengel. 2015.
\newblock \href {https://doi.org/10.1145/2766462.2767755} {{Image-Based Recommendations on Styles and Substitutes}}.
\newblock In \emph{Proceedings of the 38th International ACM SIGIR Conference on Research and Development in Information Retrieval}, pages 43--52.

\bibitem[{McAuley and Yang(2016)}]{mcauley2016addressing}
Julian McAuley and Alex Yang. 2016.
\newblock \href {https://doi.org/10.1145/2872427.2883044} {{Addressing Complex and Subjective Product-Related Queries with Customer Reviews}}.
\newblock In \emph{Proceedings of the 25th International Conference on World Wide Web}, pages 625--635.

\bibitem[{Muennighoff et~al.(2025)Muennighoff, Su, Wang, Yang, Wei, Yu, Singh, and Kiela}]{muennighoff2025gritlm}
Niklas Muennighoff, Hongjin Su, Liang Wang, Nan Yang, Furu Wei, Tao Yu, Amanpreet Singh, and Douwe Kiela. 2025.
\newblock \href {https://openreview.net/forum?id=BC4lIvfSzv} {Generative representational instruction tuning}.
\newblock In \emph{International Conference on Learning Representations}.

\bibitem[{{OpenAI}(2026{\natexlab{a}})}]{openai2026codexcli}
{OpenAI}. 2026{\natexlab{a}}.
\newblock {Codex CLI}.
\newblock \url{https://learn.chatgpt.com/docs/codex/cli}.

\bibitem[{{OpenAI}(2026{\natexlab{b}})}]{openai2026gpt6astra}
{OpenAI}. 2026{\natexlab{b}}.
\newblock {GPT-6 Astra}.
\newblock \url{https://developers.openai.com/api/docs/models/gpt-6-astra}.

\bibitem[{{OpenAI}(2026{\natexlab{c}})}]{openai2026gpt6luna}
{OpenAI}. 2026{\natexlab{c}}.
\newblock {GPT-6 Luna}.
\newblock \url{https://developers.openai.com/api/docs/models/gpt-6-luna}.

\bibitem[{Pan et~al.(2024)Pan, Luo, Wang, Chen, Wang, and Wu}]{pan2024unifying}
Shirui Pan, Linhao Luo, Yufei Wang, Chen Chen, Jiapu Wang, and Xindong Wu. 2024.
\newblock \href {https://arxiv.org/abs/2306.08302} {{Unifying Large Language Models and Knowledge Graphs: A Roadmap}}.
\newblock \emph{IEEE Transactions on Knowledge and Data Engineering}.

\bibitem[{Polonuer et~al.(2026)Polonuer, Vittor, Arango, Noori, Clifton, Del~Corro, and Zitnik}]{polonuer2026ark}
Joaquín Polonuer, Lucas Vittor, Iñaki Arango, Ayush Noori, David~A. Clifton, Luciano Del~Corro, and Marinka Zitnik. 2026.
\newblock \href {https://doi.org/10.18653/v1/2026.acl-long.714} {{Autonomous Knowledge Graph Exploration with Adaptive Breadth-Depth Retrieval}}.
\newblock In \emph{Proceedings of the 64th Annual Meeting of the Association for Computational Linguistics (Volume 1: Long Papers)}, pages 15694--15709.

\bibitem[{Qian et~al.(2023)Qian, Han, Fung, Qin, Liu, and Ji}]{qian2023creator}
Cheng Qian, Chi Han, Yi~Fung, Yujia Qin, Zhiyuan Liu, and Heng Ji. 2023.
\newblock \href {https://doi.org/10.18653/v1/2023.findings-emnlp.462} {{CREATOR}: Tool creation for disentangling abstract and concrete reasoning of large language models}.
\newblock In \emph{Findings of the Association for Computational Linguistics: EMNLP 2023}, pages 6922--6939, Singapore. Association for Computational Linguistics.

\bibitem[{Robertson and Zaragoza(2009)}]{robertson2009bm25}
Stephen Robertson and Hugo Zaragoza. 2009.
\newblock \href {https://doi.org/10.1561/1500000019} {The probabilistic relevance framework: {BM25} and beyond}.
\newblock \emph{Foundations and Trends in Information Retrieval}, 3(4):333--389.

\bibitem[{Santhanam et~al.(2022)Santhanam, Khattab, Saad-Falcon, Potts, and Zaharia}]{santhanam2022colbertv2}
Keshav Santhanam, Omar Khattab, Jon Saad-Falcon, Christopher Potts, and Matei Zaharia. 2022.
\newblock \href {https://doi.org/10.18653/v1/2022.naacl-main.272} {{ColBERTv2}: Effective and efficient retrieval via lightweight late interaction}.
\newblock In \emph{Proceedings of the 2022 Conference of the North American Chapter of the Association for Computational Linguistics: Human Language Technologies}, pages 3715--3734, Seattle, United States. Association for Computational Linguistics.

\bibitem[{Sen et~al.(2026)Sen, Kasturi, Lumer, Gulati, and Subbiah}]{sen2026grepneedagentharnesses}
Sahil Sen, Akhil Kasturi, Elias Lumer, Anmol Gulati, and Vamse~Kumar Subbiah. 2026.
\newblock \href {https://arxiv.org/abs/2605.15184} {Is grep all you need? {How} agent harnesses reshape agentic search}.
\newblock \emph{arXiv preprint arXiv:2605.15184}.

\bibitem[{Sinha et~al.(2015)Sinha, Shen, Song, Ma, Eide, Hsu, and Wang}]{sinha2015mas}
Arnab Sinha, Zhihong Shen, Yang Song, Hao Ma, Darrin Eide, Bo-June~(Paul) Hsu, and Kuansan Wang. 2015.
\newblock \href {https://doi.org/10.1145/2740908.2742839} {{An Overview of Microsoft Academic Service (MAS) and Applications}}.
\newblock In \emph{Proceedings of the 24th International Conference on World Wide Web}, pages 243--246.

\bibitem[{Sun et~al.(2024)Sun, Xu, Tang, Wang, Lin, Gong, Ni, Shum, and Guo}]{sun2024tog}
Jiashuo Sun, Chengjin Xu, Lumingyuan Tang, Saizhuo Wang, Chen Lin, Yeyun Gong, Lionel~M. Ni, Heung-Yeung Shum, and Jian Guo. 2024.
\newblock \href {https://arxiv.org/abs/2307.07697} {Think-on-graph: Deep and responsible reasoning of large language model on knowledge graph}.
\newblock In \emph{International Conference on Learning Representations}.

\bibitem[{Talmor and Berant(2018)}]{talmor2018cwq}
Alon Talmor and Jonathan Berant. 2018.
\newblock \href {https://doi.org/10.18653/v1/N18-1059} {{The Web as a Knowledge-Base for Answering Complex Questions}}.
\newblock In \emph{Proceedings of the 2018 Conference of the North {A}merican Chapter of the Association for Computational Linguistics: Human Language Technologies, Volume 1 (Long Papers)}, pages 641--651.

\bibitem[{Tam et~al.(2026)Tam, Lin, Chen, Sun, and Lee}]{tam2026smith}
Zhi~Rui Tam, Chieh-Yen Lin, Yun-Nung Chen, Shao-Hua Sun, and Hung-yi Lee. 2026.
\newblock \href {https://arxiv.org/abs/2608.24571} {Joint optimization of tool creation and use for large language model agents}.
\newblock \emph{arXiv preprint arXiv:2608.24571}.

\bibitem[{Tao et~al.(2026)Tao, Wang, Song, Luo, Liu, and Liu}]{tao2026grasp}
Yicheng Tao, Yiqun Wang, Xiangchen Song, Xin Luo, Kai Liu, and Jie Liu. 2026.
\newblock \href {https://arxiv.org/abs/2605.30237v1} {{GRASP: Plan-Guided Graph Retrieval with Adaptive Fusion and Reranking on Semi-Structured Knowledge Bases}}.
\newblock \emph{Preprint}, arXiv:2605.30237v1.

\bibitem[{Wang et~al.(2020)Wang, Shen, Huang, Wu, Dong, and Kanakia}]{wang2020mag}
Kuansan Wang, Zhihong Shen, Chiyuan Huang, Chieh-Han Wu, Yuxiao Dong, and Anshul Kanakia. 2020.
\newblock \href {https://doi.org/10.1162/qss_a_00021} {{Microsoft Academic Graph: When experts are not enough}}.
\newblock \emph{Quantitative Science Studies}, 1(1):396--413.

\bibitem[{Wang et~al.(2024{\natexlab{a}})Wang, Chen, Yuan, Zhang, Li, Peng, and Ji}]{wang2024codeact}
Xingyao Wang, Yangyi Chen, Lifan Yuan, Yizhe Zhang, Yunzhu Li, Hao Peng, and Heng Ji. 2024{\natexlab{a}}.
\newblock \href {https://arxiv.org/abs/2402.01030v4} {Executable code actions elicit better {LLM} agents}.
\newblock In \emph{Proceedings of the 41st International Conference on Machine Learning}, volume 235 of \emph{Proceedings of Machine Learning Research}, pages 50208--50232.

\bibitem[{Wang et~al.(2024{\natexlab{b}})Wang, Neubig, and Fried}]{wang2024trove}
Zhiruo Wang, Graham Neubig, and Daniel Fried. 2024{\natexlab{b}}.
\newblock \href {https://arxiv.org/abs/2401.12869v1} {{TroVE}: Inducing verifiable and efficient toolboxes for solving programmatic tasks}.
\newblock In \emph{Proceedings of the 41st International Conference on Machine Learning}, volume 235 of \emph{Proceedings of Machine Learning Research}, pages 51177--51191.

\bibitem[{W{\"o}lflein et~al.(2025)W{\"o}lflein, Ferber, Truhn, Arandjelovic, and Kather}]{wolflein2025toolmaker}
Georg W{\"o}lflein, Dyke Ferber, Daniel Truhn, Ognjen Arandjelovic, and Jakob~Nikolas Kather. 2025.
\newblock \href {https://doi.org/10.18653/v1/2025.acl-long.1266} {{LLM} agents making agent tools}.
\newblock In \emph{Proceedings of the 63rd Annual Meeting of the Association for Computational Linguistics (Volume 1: Long Papers)}, pages 26092--26130, Vienna, Austria. Association for Computational Linguistics.

\bibitem[{Wu et~al.(2024{\natexlab{a}})Wu, Zhao, Huang, Huang, Yasunaga, Cao, Ioannidis, Subbian, Leskovec, and Zou}]{wu2024avatar}
Shirley Wu, Shiyu Zhao, Qian Huang, Kexin Huang, Michihiro Yasunaga, Kaidi Cao, Vassilis~N. Ioannidis, Karthik Subbian, Jure Leskovec, and James Zou. 2024{\natexlab{a}}.
\newblock \href {https://doi.org/10.52202/079017-0817} {{AvaTaR: Optimizing LLM Agents for Tool Usage via Contrastive Reasoning}}.
\newblock In \emph{Advances in Neural Information Processing Systems}, volume~37, pages 25981--26010.

\bibitem[{Wu et~al.(2024{\natexlab{b}})Wu, Zhao, Yasunaga, Huang, Cao, Huang, Ioannidis, Subbian, Zou, and Leskovec}]{wu2024stark}
Shirley Wu, Shiyu Zhao, Michihiro Yasunaga, Kexin Huang, Kaidi Cao, Qian Huang, Vassilis~N. Ioannidis, Karthik Subbian, James Zou, and Jure Leskovec. 2024{\natexlab{b}}.
\newblock \href {https://doi.org/10.52202/079017-4037} {{STaRK: Benchmarking LLM Retrieval on Textual and Relational Knowledge Bases}}.
\newblock In \emph{Advances in Neural Information Processing Systems}, volume~37, pages 127129--127153.

\bibitem[{Xia et~al.(2025{\natexlab{a}})Xia, Wang, Yang, Wei, and Zhang}]{xia2025liveswe}
Chunqiu~Steven Xia, Zhe Wang, Yan Yang, Yuxiang Wei, and Lingming Zhang. 2025{\natexlab{a}}.
\newblock \href {https://arxiv.org/abs/2511.13646} {{Live-SWE-agent}: Can software engineering agents self-evolve on the fly?}
\newblock \emph{arXiv preprint arXiv:2511.13646}.

\bibitem[{Xia et~al.(2025{\natexlab{b}})Xia, Wu, Kim, Yu, Rossi, Wang, and McAuley}]{xia2024kar}
Yu~Xia, Junda Wu, Sungchul Kim, Tong Yu, Ryan~A. Rossi, Haoliang Wang, and Julian McAuley. 2025{\natexlab{b}}.
\newblock \href {https://doi.org/10.18653/v1/2025.naacl-long.216} {{Knowledge-Aware Query Expansion with Large Language Models for Textual and Relational Retrieval}}.
\newblock In \emph{Proceedings of the 2025 Conference of the Nations of the Americas Chapter of the Association for Computational Linguistics: Human Language Technologies (Volume 1: Long Papers)}, pages 4275--4286.

\bibitem[{Xiong et~al.(2024)Xiong, Bao, and Zhao}]{xiong2024interactivekbqa}
Guanming Xiong, Junwei Bao, and Wen Zhao. 2024.
\newblock \href {https://doi.org/10.18653/v1/2024.acl-long.569} {{Interactive-{KBQA}: Multi-Turn Interactions for Knowledge Base Question Answering with Large Language Models}}.
\newblock In \emph{Proceedings of the 62nd Annual Meeting of the Association for Computational Linguistics (Volume 1: Long Papers)}, pages 10561--10582.

\bibitem[{Xu et~al.(2024)Xu, He, Chen, Wang, Song, Tong, Liu, Zhao, and Liu}]{xu2024gog}
Yao Xu, Shizhu He, Jiabei Chen, Zihao Wang, Yangqiu Song, Hanghang Tong, Guang Liu, Jun Zhao, and Kang Liu. 2024.
\newblock \href {https://doi.org/10.18653/v1/2024.emnlp-main.1023} {{Generate-on-Graph: Treat {LLM} as both Agent and {KG} for Incomplete Knowledge Graph Question Answering}}.
\newblock In \emph{Proceedings of the 2024 Conference on Empirical Methods in Natural Language Processing}, pages 18410--18430.

\bibitem[{Yang et~al.(2025)Yang, Zhang, Jiang, and Li}]{yang2025pasemiqa}
Hansi Yang, Qi~Zhang, Wei Jiang, and Jianguo Li. 2025.
\newblock \href {https://arxiv.org/abs/2502.21087v1} {{PASemiQA: Plan-Assisted Agent for Question Answering on Semi-Structured Data with Text and Relational Information}}.
\newblock \emph{Preprint}, arXiv:2502.21087v1.

\bibitem[{Yao et~al.(2023)Yao, Zhao, Yu, Du, Shafran, Narasimhan, and Cao}]{yao2023react}
Shunyu Yao, Jeffrey Zhao, Dian Yu, Nan Du, Izhak Shafran, Karthik Narasimhan, and Yuan Cao. 2023.
\newblock \href {https://arxiv.org/abs/2210.03629} {{ReAct: Synergizing Reasoning and Acting in Language Models}}.
\newblock In \emph{International Conference on Learning Representations}.

\bibitem[{Yih et~al.(2016)Yih, Richardson, Meek, Chang, and Suh}]{yih2016webqsp}
Wen-tau Yih, Matthew Richardson, Chris Meek, Ming-Wei Chang, and Jina Suh. 2016.
\newblock \href {https://doi.org/10.18653/v1/P16-2033} {{The Value of Semantic Parse Labeling for Knowledge Base Question Answering}}.
\newblock In \emph{Proceedings of the 54th Annual Meeting of the Association for Computational Linguistics (Volume 2: Short Papers)}, pages 201--206.

\bibitem[{Yuan et~al.(2024)Yuan, Chen, Wang, Fung, Peng, and Ji}]{yuan2024craft}
Lifan Yuan, Yangyi Chen, Xingyao Wang, Yi~Fung, Hao Peng, and Heng Ji. 2024.
\newblock \href {https://arxiv.org/abs/2309.17428v2} {{CRAFT}: Customizing {LLMs} by creating and retrieving from specialized toolsets}.
\newblock In \emph{International Conference on Learning Representations}, pages 40097--40125.

\bibitem[{Zechner and {Pi Contributors}(2026)}]{zechner2026pi}
Mario Zechner and {Pi Contributors}. 2026.
\newblock Pi: A minimal terminal coding harness.
\newblock \url{https://shittycodingagent.ai}.

\bibitem[{Zhao et~al.(2025)Zhao, Zhang, Lin, Li, Wu, Zhang, and Wei}]{zhao2025pyvision}
Shitian Zhao, Haoquan Zhang, Shaoheng Lin, Ming Li, Qilong Wu, Kaipeng Zhang, and Chen Wei. 2025.
\newblock \href {https://arxiv.org/abs/2507.07998} {{PyVision}: Agentic vision with dynamic tooling}.
\newblock \emph{arXiv preprint arXiv:2507.07998}.

\end{thebibliography}

\appendix
\raggedbottom
\clubpenalty=10000
\widowpenalty=10000

\clearpage

\section{Knowledge Graph Export}
\label{app:corpus}

We export each of the three STaRK knowledge graphs once from the graph data released with the benchmark (Hugging Face dataset \texttt{snap-stanford/stark}, revision \texttt{88269e2}), using a script that makes no model calls. For Amazon, we use the default graph, which includes brand, category and color nodes. STaRK built these graphs from Amazon product reviews and metadata~\citep{he2016ups,mcauley2015image} together with Amazon question-answer data~\citep{mcauley2016addressing}, from the Microsoft Academic Graph~\citep{sinha2015mas,wang2020mag} through the OGB datasets \texttt{ogbn-mag} and \texttt{ogbn-papers100M}~\citep{hu2020ogb}, and from PrimeKG~\citep{chandak2023primekg}. Each exported graph consists of four parts.

\smallskip
\noindent \textbf{Node files.} Each node is stored as one JSON line with four fields: \texttt{node\_id} (the benchmark ID, which is also the answer ID), \texttt{node\_type}, \texttt{is\_candidate} and \texttt{attributes}. The \texttt{attributes} field contains all source attributes of the node unchanged, except that non-finite numbers are written as \texttt{null}. Nodes are stored in ID order in shards of 10{,}000 nodes each (\texttt{nodes/NNNN.jsonl}).

\smallskip
\noindent \textbf{Edge files.} The edges of each relation type are stored in one tab-separated file, \texttt{edges/RR.tsv}, where \texttt{RR} is the relation-type ID. Each file has a header row (\texttt{source\_id}, \texttt{target\_id}) and one row for every edge stored in the benchmark's edge index.

\smallskip
\noindent \textbf{Schema and candidates.} \texttt{schema.json} lists the dataset name, the node and candidate counts, the node and relation types, the number of stored rows per relation type, the shard size and short notes on the graph semantics. \texttt{candidate\_ids.txt} lists the IDs that may be returned: products (Amazon), papers (MAG) and all nodes (Prime).

\smallskip
\noindent \textbf{Format description.} \texttt{DATA\_FORMAT.txt} is identical for the three datasets. In the box below and in every other prompt or file box of the appendices, the text is reproduced exactly, line by line. A line that is too long for the box is wrapped at a space and continues on indented lines. Boxes with a dark blue title contain text that we send to the model. Boxes with a gray title contain files of the graph directory, text produced by the harness or, in Appendix~\ref{app:cases}, the request and gold answers of a case. Boxes with a green or red title contain trajectory excerpts (Appendix~\ref{app:cases}). \texttt{DATA\_FORMAT.txt} reads:
\begin{dkgifile}{\texttt{DATA\_FORMAT.txt} (Amazon, MAG, Prime)}
Node records: nodes/NNNN.jsonl, shard number = node_id // 10000.
Each line has node_id (benchmark ID), node_type, is_candidate, and full attributes.
IDs inside attributes can be external source IDs; return node_id in predictions.
Graph edges: edges/RR.tsv, source_id and target_id columns with a header.
RR is the relation type ID from schema.json. Stored edge rows are not truncated.
For adjacency, inspect either endpoint; the official default graph is undirected.
Preserve relation meaning when interpreting directions or entity roles.
Only IDs in candidate_ids.txt are valid retrieval outputs. Other nodes are usable evidence.
Use shell tools or Python's json/csv modules to read these files.
\end{dkgifile}
\noindent The prompts tell the agent to read this file and \texttt{schema.json} (Appendix~\ref{app:prompt}). \vshell{} and the runs with the rg prompt omit the last line of this file, and \vnoedges{} omits the four lines about edges (Appendices~\ref{app:prompt-texts} and~\ref{app:harness}).

\smallskip
\noindent \textbf{What the export preserves.} The exported graphs keep all nodes, including the evidence nodes outside the candidate set (for example, brands in Amazon and authors in MAG), with all their attributes, every stored edge row and the type labels. The export is a plain-text copy of the released graph, with the node types, relation types and attribute fields of the release, and predictions are scored directly on the benchmark IDs. After exporting, the script checks that the node files contain every node exactly once and in ID order, and that 100 randomly chosen records per dataset read back identical to the source. Queries and gold answers are kept in files outside the agent's sandbox. Table~\ref{tab:corpus} gives the sizes of the exported graphs.

\begin{table}[t]
\centering
\small
\setlength{\tabcolsep}{4pt}
\begin{tabular}{@{}lrrr@{}}
\toprule
 & Amazon & MAG & Prime \\
\midrule
Node types & 4 & 4 & 10 \\
Relation types & 5 & 4 & 18 \\
Nodes & 1{,}035{,}542 & 1{,}872{,}968 & 129{,}375 \\
Candidates & 957{,}192 & 700{,}244 & 129{,}375 \\
Stored edge rows & 5{,}382{,}103 & 19{,}919{,}698 & 8{,}100{,}498 \\
\midrule
Node shards & 104 & 188 & 13 \\
Node files (GB) & 6.47 & 1.87 & 0.12 \\
\quad per node (KB) & 6.2 & 1.0 & 0.9 \\
Edge files & 5 & 4 & 18 \\
Edge files (GB) & 0.076 & 0.311 & 0.093 \\
Export total (GB) & 6.55 & 2.18 & 0.21 \\
\bottomrule
\end{tabular}
\caption{Size of the exported graphs (1\,GB = $10^9$ bytes, 1\,KB = $10^3$ bytes). The row ``per node'' gives the size of the node files divided by the number of nodes. The export total includes \texttt{schema.json}, \texttt{DATA\_FORMAT.txt} and \texttt{candidate\_ids.txt}. STaRK reports 9{,}443{,}802, 39{,}802{,}116 and 8{,}100{,}498 relations for Amazon, MAG and Prime, respectively. For Amazon and MAG, these counts equal the number of distinct ordered node pairs after every stored row is also added in reverse.}
\label{tab:corpus}
\end{table}

\section{Prompts and Prompt Validation}
\label{app:prompt}

The prompts we compared differ in two parts: the ranking instruction, which specifies how many IDs to return and how to rank them, and the tool instruction, which suggests tools for the agent to use. Appendix~\ref{app:prompt-texts} gives the texts and Appendix~\ref{app:validation} the validation runs that compared them.

\subsection{Prompt Texts}
\label{app:prompt-texts}

Every run sends the task as a single user message and leaves the harness's system prompt unchanged (Appendix~\ref{app:harness}). Placeholders are shown in angle brackets: \dkgiph{query} is the query text, \dkgiph{corpus dir} the directory of the exported graph (Appendix~\ref{app:corpus}), \dkgiph{scratch dir} a writable directory created for each query, and \dkgiph{attempt dir} the run directory of the query, which contains the scratch directory. The ranking instruction is marked with \outpolmark{} (blue text) and the tool instruction with \toolhintmark{} (orange text). The marks are annotations added for this paper.

\smallskip
\noindent \textbf{Final prompt.} All test runs of \vfull{} use the final prompt, shown at the top of page~\pageref{box:final}. Its ranking instruction asks the agent to check its top candidate against every requirement of the query and then to rank further plausible candidates after it, returning at most 20 IDs.
\begin{figure*}[t]
\begin{dkgiprompt}{Final prompt (\vfull{})}
Answer the retrieval query using only the supplied local STaRK knowledge base. Use the textual attributes and graph relations as evidence. Return candidate node IDs ranked by relevance, most relevant first. Use the benchmark node_id field, not external IDs within attributes. Only nodes marked is_candidate=true are valid answers; other node types may be used as evidence. @<The query may have several correct answers. Return 20 unique integer IDs observed in the corpus, ranked by how well the evidence supports them. Before ranking a candidate first, check it against every requirement in the query; entities that the query mentions only as a reference point are not answers unless the query asks for them. After the first candidate, add the other plausible candidates you found in order of support; they do not each need the same level of verification. Return fewer than 20 only if you cannot find that many plausible candidates. Do not invent IDs.>@

Return only JSON with exactly this structure: {"node_ids": [123, 456]}. Do not include Markdown, explanations, or citations outside the JSON.

Query: <query>

Corpus directory: <corpus dir>
Read DATA_FORMAT.txt and schema.json for the file format, entity types, relation types, and graph interpretation. Nodes and edges are complete local text files. @|Use shell tools, rg, or Python 3 standard-library modules to search and read them.|@ The corpus is read-only. Writable scratch directory: <scratch dir>
\end{dkgiprompt}
\phantomsection\label{box:final}
\end{figure*}

\smallskip
\noindent \textbf{rg prompt.} The rg prompt, used in the tool-instruction comparison of Appendix~\ref{app:validation}, replaces our tool instruction with that of \citet{li2026beyond}, given in the first box below. Its runs also omit the last line of \texttt{DATA\_FORMAT.txt} (\emph{``Use shell tools or Python's json/csv modules to read these files.''}). This variant removes every mention of Python from the prompt and the format description. Both prompts of this comparison use the earlier ranking instruction in the second box below in place of the final one, and the rest of the rg prompt is identical to the final prompt.
\begin{dkgiprompt}{Tool instruction of the rg prompt}
Use ripgrep (rg) instead of grep for fast searching.
\end{dkgiprompt}
\begin{dkgiprompt}{Earlier ranking instruction (tool-instruction comparison)}
@<The query may have several correct answers. Return 20 unique integer IDs observed in the corpus, ranked by how well the evidence supports them: first the candidates you are most confident in, then the next most plausible candidates you found. Return fewer than 20 only if you cannot find that many plausible candidates. Do not invent IDs.>@
\end{dkgiprompt}

\smallskip
\noindent \textbf{Ablation variants.} Each variant of \S\ref{sec:ablation} keeps the final prompt verbatim up to and including the query line and changes the tool section, which in the final prompt runs from \emph{``Corpus directory:''} to the end. \vcodex{} uses the final prompt unchanged. In \vshell{}, the tool instruction becomes \emph{``Use shell tools and rg to search and read them.''}, and the line in the first box below is added at the end. \vone{} keeps the final prompt and adds the line in the second box at the end.
\begin{dkgiprompt}{Line added at the end of the prompt of \vshell{}}
General-purpose programming languages are unavailable; use only shell text tools and bash syntax.
\end{dkgiprompt}
\begin{dkgiprompt}{Line added at the end of the prompt of \vone{}}
Python 3 can be started only once in this task; after that, python3 is unavailable and only shell text tools and bash remain.
\end{dkgiprompt}
\noindent \vnoedges{} replaces \emph{``Nodes and edges are complete local text files.''} with \emph{``Nodes are complete local text files.''} and leaves the task text unchanged, although it mentions graph relations. The variants with predefined graph tools replace the whole tool section with the texts below. \vretr{} removes the signature of \texttt{neighbors} and the two sentences about \texttt{neighbors} and its optional filters from the tool section of \vtools{}.
\begin{dkgiprompt}{Tool section of \vtools{}}
Available tools: get_schema(), search(query, k=5), get_node(node_id), neighbors(node_id, relation, node_type, query, k=20). Call get_schema first to understand node and relation types. Search returns up to 30 BM25 matches with benchmark IDs, names, types, and summary context. get_node returns the complete record including attributes and is_candidate. neighbors examines one hop in either direction, optionally filters relation and node type, and uses keyword matching when query is supplied; k is at most 20. Optional filters may be omitted or empty. Use node_ids from returned records.
\end{dkgiprompt}

\begin{dkgiprompt}{Tool section of \vtoolscode{}}
Available tools: read and bash, including Python 3. Knowledge-base files cannot be read directly. In Python, use: import sys; sys.path.insert(0, '<attempt dir>'); import kbtools; kb = kbtools.Client('<attempt dir>/kb.sock'). Call kb.get_schema() first to understand node and relation types. kb.search(query, k=5) returns up to 30 BM25 matches. kb.get_node(node_id) returns the complete record including attributes and is_candidate. kb.neighbors(node_id, relation='', node_type='', query='', k=20) examines one hop in either direction with optional filters and keyword matching; k is at most 20. All methods return text. Writable scratch directory: <scratch dir>
\end{dkgiprompt}

\smallskip
\noindent \textbf{Closed-book probes.} The two probes (Appendix~\ref{app:probes}) are closed-book. The prompt of the ID probe is:
\begin{dkgiprompt}{Closed-book ID probe}
Answer the retrieval query from your own knowledge. You do not have access to the knowledge base, its files, or any tools.
Knowledge base: <KB description>. Valid answers: <answer types>.
The query may have several correct answers.
Return the STaRK benchmark node_id integers of the correct answers. If you remember them, list them; otherwise give your best guesses.
Return 20 unique integer IDs ranked by confidence, most confident first. Return fewer than 20 only if you cannot give that many.
Return only JSON with exactly this structure: {"node_ids": [123, 456]}. Do not include Markdown, explanations, or citations outside the JSON.

Query: <query>
\end{dkgiprompt}
\noindent The name probe, run on MAG and Prime, replaces the three lines after \emph{``The query may have several correct answers.''} with three lines that ask for names: \emph{``Return the \textup{\dkgiph{name kind}} of the correct answers.''}, the same ranking line with \emph{``20 unique names''} in place of \emph{``20 unique integer IDs''}, and the same JSON line with the structure \texttt{\{"names":~["...",~"..."]\}}. \dkgiph{KB description} names the knowledge base (for example, \emph{``the STaRK-MAG academic knowledge base built from the Microsoft Academic Graph''}), and \dkgiph{answer types} names the candidate types: products, papers, or the ten node types of Prime. \dkgiph{name kind} is \emph{``exact paper titles''} for MAG and \emph{``exact entity names as used in PrimeKG''} for Prime.

\subsection{Prompt Validation}
\label{app:validation}

All prompt decisions were made on STaRK validation queries with GPT-6-Luna in both harnesses, without using any test query, test trajectory or test label. A prompt-selection round compared ranking instructions and tool instructions. In its tool-instruction comparison on the same 450 queries (150 per dataset) for both prompts, the rg prompt reduced the share of shell commands that run Python from 47\% to 5\% with Codex and from 51\% to 1.5\% with Pi, relative to the prompt with the same ranking instruction and our tool instruction. With Codex, the share of queries in which the agent ran any Python fell from 98\% to 12\%. Input tokens rose by about 110k per query with both harnesses. We then revised the ranking instruction once, which produced the final prompt. On 750 further validation queries (300 from Amazon, 300 from MAG and 150 from Prime), we compared it with a base prompt whose ranking instruction allows up to 100 IDs and whose other text is the same. Overall \rtwenty{} rose from 61.9 to 71.3 with Pi and from 57.6 to 68.8 with Codex, and overall \hone{} changed by $-2.3$ and $-0.8$ points, respectively. The prompt was fixed before any test result was seen.

\section{Harness Details}
\label{app:harness}

All runs of a configuration use the same harness, and each query has its own scratch directory. The agent writes its programs inline in its commands, and they are discarded after each query. All runs that we conducted, including the validation runs and the closed-book probes, used GPT-6-Luna or GPT-6-Astra through an API in September and October 2026.

\smallskip
\noindent \textbf{Pi.} We use DCI-Agent-Lite~\citep{li2026beyond} at upstream commit \texttt{271f37e} (1 June 2026). DCI-Agent-Lite builds on the Pi coding agent~\citep{zechner2026pi},\footnote{\url{https://github.com/badlogic/pi-mono}} using a fork of Pi version 0.65.0 to which it adds runtime context management. We use both as released. Our three small changes to DCI-Agent-Lite's runner script are in our code release. The one that affects scoring reads the final answer from the last assistant message, so that an attempt that ends with a tool call or an error counts as failed. Pi runs in its offline mode with two tools, \texttt{read} and \texttt{bash}, and with its default system prompt, which lists the available tools and gives the date and the working directory. Its skills, prompt templates, themes and extensions are disabled, except for the extension that registers the graph tools. The model is GPT-6-Luna (GPT-6-Astra for the stronger-backbone rows) with reasoning effort \emph{high}, a context window of 272{,}000 tokens and at most 32{,}000 output tokens per request. For context management, we use level~3 of DCI-Agent-Lite, the lightweight setting that it uses for its own benchmarks, which truncates large tool outputs and compacts the history by replacing older tool results with a placeholder. Each attempt may take at most 300 turns and 1{,}800\,s of wall-clock time. Head-and-tail truncation, described below, is our one change to how Pi returns tool outputs.

\smallskip
\noindent \textbf{Codex CLI.} \vcodex{} runs the \texttt{codex exec} command of Codex CLI 0.156.1\footnote{\url{https://github.com/openai/codex}} with the same model, reasoning effort and context window as Pi, and its budget is Pi's time limit per attempt. User configuration, rules and project documents are ignored, sessions are ephemeral, and approval is never requested. Web search, network access and optional features such as multi-agent use, memory, plugins and browser use are disabled. Codex's own system prompt and tool are used unchanged. With this tool, the model writes short JavaScript programs, each of which can run several shell commands in one call. Codex keeps its own output truncation, which lets the model raise the limit for an individual call, and we apply head-and-tail truncation in Pi.

\smallskip
\noindent \textbf{Sandbox.} Both harnesses run in the same outer sandbox, a deny-by-default Landlock policy under which system libraries, the harness code and the working directory are read-only. The working directory is the graph directory or, for the graph-tool variants, a directory that holds a single README file. The writable paths are an installation-ID file of Codex and the run directory of the query, which holds the scratch directory, the logs and a per-query home directory. In Pi, shell commands run in a second sandbox layer with a clean environment, a second Landlock policy and a seccomp filter that blocks network sockets. In Codex, they run in Codex's own sandbox with network access disabled. Only the harness process itself connects to the model endpoint, and no variant has a web-search tool. Before the model is first called, a check inside the sandbox confirms that the working directory is read-only and that the query and label files, the original benchmark files and the user's home configuration are unreadable. The scoring process reads the labels after the agent has finished.

\smallskip
\noindent \textbf{Head-and-tail truncation.} In every Pi variant with a shell, Pi's \texttt{bash} tool runs its commands through a wrapper that implements the head-and-tail truncation of \S\ref{sec:outputs}. The wrapper saves the complete output of every command in \dkgiph{scratch dir}\texttt{/full\_outputs/}, where the agent's later commands can read it. An output of at most 18{,}000 bytes and 1{,}800 lines is returned as it is. For a longer output, the wrapper returns the first 9{,}000 bytes or 900 lines, whichever limit is reached first, then a notice on a separate line, and then the last 9{,}000 bytes or 900 lines. The notice reads:
\begin{dkgifile}{Truncation notice of the \texttt{bash} wrapper}
[... harness output limit: showing the first <h> lines (<hb> bytes) and the last <t> lines (<tb> bytes) of <n> lines (<nb> bytes); <ob> bytes omitted in between. Complete output saved to <path> - inspect it with rg, sed -n, head or tail. ...]
\end{dkgifile}
\noindent The command's exit status is preserved. A truncated output stays below the 50\,KB limit of Pi's \texttt{bash} tool and the 20{,}000-character limit of DCI-Agent-Lite's context management, so the model receives both of its ends together with the notice. The \texttt{read} tool keeps Pi's limits (2{,}000 lines or 50\,KB per call) and the 20{,}000-character limit of context management.

\smallskip
\noindent \textbf{Graph tools.} \vtools{}, \vretr{} and \vtoolscode{} access the graph through the tools in Table~\ref{tab:graph-tools}. The tools are based on those of ARK~\citep{polonuer2026ark}. The \texttt{search} and \texttt{neighbors} tools call ARK's released implementations of its global search and one-hop neighborhood search, which run on ARK's prepared graph (BM25 over ARK's node names and summaries), and \texttt{get\_node} and \texttt{get\_schema} return a node record and the schema file of our graph export. The agent, the prompt and the step budget come from our harness. The tools are served by a process outside the sandbox, and the graph files also stay outside it.
In \vtools{} and \vretr{}, the tools are registered in Pi as native function-calling tools with the descriptions in Table~\ref{tab:graph-tools}, and Pi's built-in tools are disabled, so every action of the agent is a graph-tool call. In \vtoolscode{}, the agent has Pi's \texttt{read} and \texttt{bash} tools and calls the same four tools from Python through a client whose methods take the same arguments and return the same text. On the test subset, \vtoolscode{} made an average of 45.1 graph-tool calls per query within 6.8 \texttt{bash} calls, compared with 18.4 tool calls per query for \vtools{}.

\begin{table}[t]
\centering
\footnotesize
\begin{tabular}{@{}p{\linewidth}@{}}
\toprule
\raggedright\texttt{get\_schema()}\\
\quad Returns \texttt{schema.json}. \tabularnewline
\midrule
\raggedright\texttt{search(query, k=5)}\\
\quad The $k\le30$ best BM25 matches for \texttt{query} over all nodes, each with ID, name, type, score and a short context around the first matching query word. \tabularnewline
\midrule
\raggedright\texttt{get\_node(node\_id)}\\
\quad The complete JSON record of the node, as stored in the graph export. \tabularnewline
\midrule
\raggedright\texttt{neighbors(node\_id, relation=\textquotesingle\textquotesingle, node\_type=\textquotesingle\textquotesingle, query=\textquotesingle\textquotesingle, k=20)}\\
\quad The one-hop neighbors in either direction, optionally restricted to one relation and one node type. With \texttt{query}, the tool keeps the neighbors whose name contains the query or whose summary contains query words and orders them with name matches first and then by the number of matched words. Shows the first $k\le20$ with their relation labels and the total number found. \tabularnewline
\bottomrule
\end{tabular}
\begin{dkgiprompt}{Tool descriptions passed to the model}
Read knowledge-base entity types, relation types, and graph interpretation.
Global ARK BM25 search; returns node IDs, names, types, and summary context.
Read a complete node record, including attributes and is_candidate.
ARK one-hop neighborhood search with optional relation, node type, and keyword filters.
\end{dkgiprompt}
\caption{Graph tools available to \vtools{}, \vretr{} and \vtoolscode{}. The box gives the description of each tool that the model receives in \vtools{} and \vretr{}, in the order of the rows. The tool section of the prompt is given in Appendix~\ref{app:prompt-texts}. \vretr{} has all tools except \texttt{neighbors}.}
\label{tab:graph-tools}
\end{table}

\smallskip
\noindent \textbf{\vshell{}.} The shell can start a fixed list of 15 programs (\texttt{rg}, \texttt{cat}, \texttt{sed}, \texttt{head}, \texttt{tail}, \texttt{wc}, \texttt{cut}, \texttt{sort}, \texttt{comm}, \texttt{ls}, \texttt{find}, \texttt{pwd}, \texttt{xargs}, \texttt{echo} and \texttt{bash}). A Landlock execute rule enforces this list, also for \texttt{find -exec} and \texttt{xargs}. Before each query, a check confirms that these programs run and that starting \texttt{python3}, \texttt{awk}, \texttt{perl} or \texttt{node} fails, testing \texttt{python3} also through its absolute path, \texttt{find -exec}, \texttt{xargs} and the dynamic loader. Pipelines, bash loops, arrays and arithmetic remain available, and \texttt{grep}, \texttt{awk} and \texttt{jq} are unavailable. The graph directory is a hard-link copy of the graph export, and its \texttt{DATA\_FORMAT.txt} omits the last line. Pi's \texttt{read} tool and head-and-tail truncation are as in \vfull{}.

\smallskip
\noindent \textbf{\vone{}.} The shell can start the 15 programs of \vshell{} under the same Landlock rules, as well as \texttt{python3}, which has a budget of one start per attempt. The \texttt{python3} command on the agent's path is a launcher that requests each start from the process that runs the query, which keeps the count outside the sandbox. Every further start, including one from within the running program, is refused with exit status 126 and the fixed message \texttt{python3: unavailable: Python 3 can be started only once in this task.} A check before each query confirms this behavior, and the count is then reset. \vone{} differs from \vfull{} in the single Python invocation, one added prompt line (Appendix~\ref{app:prompt-texts}) and the programs that the shell can start, and is identical in all other respects. On the test subset, \vfull{} ran \texttt{awk} in 21 of its 5{,}866 shell commands, spread over 14 queries, and all its other interpreter runs were Python. With \vone{}, 573 of the 710 queries used their single Python invocation.

\smallskip
\noindent \textbf{\vnoedges{}.} The graph directory holds hard links to the node shards and to \texttt{candidate\_ids.txt}, a \texttt{schema.json} from which the relation types, the edge counts and the note on graph semantics are removed, and a \texttt{DATA\_FORMAT.txt} shortened by its four lines about edges. Node records are unchanged, so information that the source also stores as attribute text (for example, an Amazon product's brand and category names) is still visible. Of the graph data, the sandboxes of \vshell{} and \vnoedges{} can each read their own graph directory.

\smallskip
\noindent \textbf{Scoring.} The agent's final message must be a JSON object with the single key \texttt{node\_ids} and a list of at most 100 entries, and any other output counts as a failed attempt. Entries are scored in order, and an entry that is a non-integer, a non-candidate ID or a repeat keeps its position and counts as a miss. \hone{} is 1 if the first entry is a gold answer. \rtwenty{} is the number of gold answers among the first 20 entries divided by the number of gold answers. MRR is the reciprocal rank of the first gold answer among the first 100 entries, and a list whose entries are all incorrect counts as a miss. Five runs on the test sets returned 21 IDs, one more than the 20 requested, and each of them has an incorrect answer at rank 21, so every reported MRR equals MRR over the first 20 entries. An attempt that ended in a provider error, at the time limit or with output other than the requested JSON was retried, and every query of a test configuration needed at most three attempts. Metrics are averaged over all queries of a set, and a query whose attempts all failed counts as a miss in every metric. In the test configurations, this occurred for one Amazon query of \vone{}, whose three attempts all reached the time limit.

\smallskip
\noindent \textbf{Cost.} Table~\ref{tab:cost} gives the cost and resource use of each configuration at official API prices. Per million input, cached input and output tokens, GPT-6-Luna costs \$0.10, \$0.01 and \$0.50, respectively, and GPT-6-Astra costs \$10, \$1 and \$50. The full test run of \vfull{} with GPT-6-Astra cost about \$3{,}000 in total. Costs are computed from the recorded token usage, which is missing for some failed attempts.

\begin{table}[t]
\centering
\footnotesize
\setlength{\tabcolsep}{3pt}
\begin{tabular}{@{}lrrrrr@{}}
\toprule
 & \$ per & & \multicolumn{2}{c}{Tokens} & Time \\
\cmidrule(lr){4-5}
Configuration & query & Calls & Input & Output & (s) \\
\midrule
\multicolumn{6}{@{}l}{\emph{Full test set (7{,}108 queries), GPT-6-Luna}} \\
\vfull{} & 0.0050 & 10 & 120{,}346 & 2{,}212 & 273 \\
\vcodex{} & 0.0082 & 10 & 250{,}080 & 3{,}930 & 170 \\
\multicolumn{6}{@{}l}{\emph{Full test set (7{,}108 queries), GPT-6-Astra}} \\
\vfull{} & 0.42 & 9 & 119{,}106 & 1{,}151 & 139 \\
\midrule
\multicolumn{6}{@{}l}{\emph{Test subset (710 queries), GPT-6-Luna}} \\
\vfull{} & 0.0050 & 10 & 119{,}067 & 2{,}170 & 272 \\
\vtoolscode{} & 0.0038 & 6 & 78{,}162 & 1{,}932 & 183 \\
\vtools{} & 0.0033 & 17 & 81{,}542 & 1{,}147 & 196 \\
\vshell{} & 0.0076 & 14 & 205{,}260 & 2{,}380 & 304 \\
\vone{} & 0.0068 & 12 & 158{,}224 & 2{,}454 & 725 \\
\vretr{} & 0.0036 & 16.5 & 89{,}574 & 1{,}254 & 211 \\
\vnoedges{} & 0.0056 & 11 & 134{,}263 & 2{,}384 & 287 \\
\vcodex{} & 0.0084 & 11 & 264{,}034 & 3{,}832 & 236 \\
\midrule
\multicolumn{6}{@{}l}{\emph{Test subset (710 queries), GPT-6-Astra}} \\
\vfull{} & 0.40 & 9 & 119{,}588 & 1{,}142 & 164 \\
\vcodex{} & 0.62 & 10 & 233{,}042 & 1{,}484 & 154 \\
\bottomrule
\end{tabular}
\caption{Cost and resource use per query. \$ per query is the mean over queries, and the other columns are medians. Calls are the agent's actions $a_t$ (\S\ref{sec:method}), input tokens include cached input, and time is wall-clock time. For \vtoolscode{}, calls are \texttt{bash} calls, each of which may call several graph tools, and for \vcodex{} they are command executions. The wall-clock time of \vone{} mainly reflects slower model responses during its run, with a median of 87\,s per model request compared with 33\,s for \vfull{}. Its single Python program ran for a median of 10\,s.}
\label{tab:cost}
\end{table}

\section{Harness Comparison on the Test Subset}
\label{app:ablation}

Table~\ref{tab:scaffolds} gives the harness comparison of \S\ref{sec:results} on the test subset. Its GPT-6-Luna rows are the \vfull{} and \vcodex{} rows of Table~\ref{tab:ablation}. The test-subset results of \vfull{} come from the runs on the full test set, and those of Codex with GPT-6-Astra from a separate run on the test subset.

\begin{table}[t]
\centering
\footnotesize
\setlength{\tabcolsep}{2.4pt}
\begin{tabular}{@{}lcccccc@{}}
\toprule
 & \multicolumn{4}{c}{\hone{}} & \multicolumn{2}{c}{Overall} \\
\cmidrule(lr){2-5}\cmidrule(l){6-7}
 & Amazon & MAG & Prime & Overall & \rtwenty{} & MRR \\
\midrule
Pi, Luna & 59.1 & 91.0 & 69.6 & 75.2 & 78.0 & 78.9 \\
Codex, Luna & 58.5 & 87.6 & 67.5 & 73.0 & 75.5 & 76.9 \\
\midrule
Pi, Astra & 72.6 & 93.2 & 79.6 & 83.1 & 84.9 & 85.9 \\
Codex, Astra & 72.6 & 93.2 & 79.6 & 83.1 & 84.8 & 86.2 \\
\bottomrule
\end{tabular}
\caption{\vfull{} on the test subset (710 queries). Each row is labeled with the harness and the model, where Luna and Astra denote GPT-6-Luna and GPT-6-Astra. All rows use the same prompt and time limit per attempt.}
\label{tab:scaffolds}
\end{table}

\section{Compared Baselines}
\label{app:baselines}

We describe the baselines of Table~\ref{tab:main} in the order and grouping of the table. For each system, we state how it accesses the KG, what it trains or tunes on STaRK, which backbone models its paper names and which configuration we report.

\subsection{Trained or Tuned on STaRK}
\noindent \textbf{AvaTaR}~\citep{wu2024avatar} is an LLM agent that scores candidate nodes with a Python program. The program is built from a library of predefined knowledge-base functions, which retrieve documents, embeddings and relations, compute similarity scores or ask an LLM to check a requirement. During optimization, a comparator LLM contrasts training queries on which the agent performs well with those on which it performs poorly and writes new instructions for the agent. The optimized program is applied directly to the test queries. We report its main results with Claude~3 Opus from Table~2 of the paper.

\smallskip
\noindent \textbf{PASemiQA}~\citep{yang2025pasemiqa} separates planning from traversal. Its Llama~2-7B planner is fine-tuned to predict the shortest relation paths from the topic nodes of a query to its answers. A GPT-4 agent, guided by the predicted paths, takes actions that expand a node to its most relevant neighbors, retrieve nodes by embedding similarity and return the answer. We report its \hone{} and MRR from Tables~2 and~10 of the paper.

\smallskip
\noindent \textbf{mFAR}~\citep{li2024mfar} is a trained retriever that represents each node by one field per property or relation. On MAG and Prime, these fields also include entities up to two hops away. mFAR scores every field with BM25 and a Contriever encoder and combines the scores with query-conditioned softmax weights. The encoder and the weights are trained on the STaRK training queries. The All+2 configuration scores both a single-document and a multi-field representation. We report it from Table~1 of the paper.

\smallskip
\noindent \textbf{MoR}~\citep{lei2025mor} mixes structural and textual retrieval. A Llama~3.2-3B planner, fine-tuned on 1{,}000 sampled training queries with LLM-generated planning graphs, outlines the entities that a query involves and the relations between them. A layer-wise traversal follows this plan and at each layer merges the neighbors of the planned node type with textual matches, which come from BM25 on Amazon and MAG and from a fine-tuned Contriever on Prime. A trained reranker then scores each candidate from features of the trajectory that reached it (the node types along the path, whether each step was structural or textual and the text similarity of each node) and from the candidate's ada-002 similarity to the query. We report the MoR row from Table~1 of the paper.

\smallskip
\noindent \textbf{GraphRAFT}~\citep{clemedtson2025graphraft} stores the KG in a Neo4j graph database and retrieves answers in three stages. GPT-4o-mini identifies the entities in the query, and a name without an exact match is resolved to its most similar nodes through a vector index over ada-002 node embeddings. A text-to-Cypher model then generates Cypher queries whose decoding is restricted to the schema-valid multi-hop queries around the identified entities, and the most probable queries are executed until the retrieved subgraph reaches a size limit. A Llama~3.1-8B-Instruct model then reads the textualized subgraph and selects and orders the answers. The text-to-Cypher model starts from Neo4j's Gemma~2-9B text2cypher checkpoint, and both models are fine-tuned with LoRA on MAG and Prime. We report the full system from Table~1 of the paper, which covers these two datasets.

\smallskip
\noindent \textbf{GRASP}~\citep{tao2026grasp} is a three-stage retrieval pipeline. A prompted Qwen3-27B-Instruct planner writes a JSON plan of anchors, hops and a target, which is compiled into one Cypher query on Neo4j. The resulting candidates are rescored with Qwen3-Embedding-4B on Amazon and MAG and with SapBERT on Prime. Their ranking is then fused with that of a trained mFAR retriever under a weight that can depend on the plan. Finally, a Qwen3-Reranker-0.6B, fine-tuned with LoRA on the STaRK training queries, reranks the top 100 candidates of the fused ranking, and its output is fused with that ranking once more. The fusion weights are tuned on the validation queries, and the Prime planner prompt contains 19 STaRK training queries with hand-curated plans. We report the full system, listed as supervised GRASP in Table~1 of the paper.

\subsection{Training-Free} 

\noindent \textbf{BM25}~\citep{robertson2009bm25}, \textbf{ColBERTv2}~\citep{santhanam2022colbertv2} and \textbf{GritLM-7B}~\citep{muennighoff2025gritlm} are lexical, late-interaction and LLM-based embedding retrievers that \citet{wu2024stark} evaluate on the candidate node documents of each KG. BM25 scores a document by the frequency of the query terms in it, adjusted for term rarity and document length, and ColBERTv2 matches token-level embeddings of the query against those of each document. GritLM-7B embeds the query and the documents with a 7B LLM trained for both embedding and generation and ranks the documents by similarity. We report their results on the full test sets from Table~6 of the STaRK paper.

\smallskip
\noindent \textbf{KAR}~\citep{xia2024kar} expands the query with an LLM before dense retrieval. It retrieves the documents of the entities mentioned in the query and collects their neighbors up to two hops away. For each entity, it keeps the ten neighbors whose documents are most similar to the query. GPT-4o writes query expansions from the resulting document triples, and ada-002 embeddings match the expanded query against the node documents. We report this configuration from Table~2 of the paper.

\smallskip
\noindent \textbf{HybGRAG}~\citep{lee2024hybgrag} is an LLM agent that routes each query to a text retriever or a hybrid retriever and refines the routing with a critic. An LLM router identifies the topic entities and the relevant relations. The hybrid retriever then follows these relations within the ego-graphs of the entities and ranks the documents it reaches by vector similarity to the query. An LLM validator checks the top results, and an LLM commenter returns corrective feedback to the router for at most four iterations. The examples in its prompts come from the STaRK training set. We report its main configuration with Claude~3 Sonnet and ada-002 embeddings from Table~5 of the paper, which covers MAG and Prime.

\smallskip
\noindent \textbf{AF-Retriever}~\citep{boer2025afretriever} is an eight-step LLM pipeline that combines graph constraints with vector similarity search. The LLM writes a Cypher query. On Prime, where answers can have several node types, it first predicts the target node type. The query is parsed into relational triples and attribute constraints, which are grounded on the graph by embedding search and set intersection. Vector search fills about two thirds of the 20 final candidates from the grounded set and the rest from unconstrained candidates of the target type, and the LLM reranks these 20 pairwise. We report its zero-shot configuration with gpt-oss-120B and text-embedding-3-small. Its \hone{} and MRR come from Table~3 of the paper, and its \rtwenty{} comes from the complete-pipeline row of the ablation table (Table~5).

\smallskip
\noindent \textbf{ARK}~\citep{polonuer2026ark} is an LLM agent that calls two predefined tools. A global search ranks node texts with BM25, and a one-hop neighborhood search can filter the neighbors of a node by node and edge type and rank them by a subquery. Three agents run in parallel for at most 20 steps each, and the nodes they select are fused by vote count. We report its GPT-5.2 results from Table~5 in the appendix of the paper. Our graph tools call ARK's implementations of these two tools (Appendix~\ref{app:harness}).

\subsection{KG Reasoning Methods Adapted to STaRK} \citet{yang2025pasemiqa} adapt three KG reasoning methods to STaRK and run them with GPT-4. We report their \hone{} and MRR from Tables~2 and~10 of that paper.

\smallskip
\noindent \textbf{ToG}~\citep{sun2024tog} runs an LLM-guided beam search on the KG that alternates relation and entity exploration and stops once the LLM judges the retained reasoning paths sufficient or the search reaches a maximum depth. 

\smallskip
\noindent \textbf{RoG}~\citep{luo2024rog} generates relation paths grounded in the KG as plans, retrieves the reasoning paths that follow these plans and lets the LLM reason over them. In its original form, RoG's planner is Llama~2-Chat-7B instruction-tuned on WebQSP and CWQ~\citep{yih2016webqsp,talmor2018cwq}.

\smallskip
\noindent \textbf{GoG}~\citep{xu2024gog} is a training-free KG agent that interleaves thinking, searching and generation steps, in which the LLM proposes new triples from its own knowledge and verifies them.

\section{Closed-Book Probes}
\label{app:probes}

The closed-book probes measure how many queries the model answers from its parametric knowledge alone (\S\ref{sec:ablation}).

\smallskip
\noindent \textbf{Design.} Both probes run GPT-6-Luna in Pi with reasoning effort \emph{high}, and the model answers in a single closed-book turn (prompts in Appendix~\ref{app:prompt}). The \emph{ID probe} asks for up to 20 STaRK node IDs, ranked by confidence, and is run on the full test sets of all three datasets (7{,}108 queries). Since node IDs are arbitrary integers, a correct ID could indicate that the model has memorized benchmark files. The \emph{name probe} asks for up to 20 answer names (exact paper titles for MAG and exact PrimeKG entity names for Prime), which we map to node IDs offline. Because Amazon product titles are long and rarely match exactly, it is run on the full MAG and Prime test sets (5{,}466 queries). The 12{,}574 probe runs cost \$8.31 in total, and the 4 runs that failed count as wrong.

\smallskip
\noindent \textbf{Name mapping.} The mapping uses the exported node files. MAG papers are indexed under their \texttt{title} and \texttt{OriginalTitle}, and Prime nodes under their \texttt{name}, \texttt{details.name} and every entry of \texttt{details.alias}. After the same normalization of indexed and returned names (NFKC, case folding, non-alphanumeric characters replaced with spaces), a returned name maps to every node indexed under the same normalized name. A query counts as answered at rank $r$ if the $r$-th returned name maps to at least one gold node.

\smallskip
\noindent \textbf{Results.} The ID probe answers 0 of 1{,}642 Amazon, 0 of 2{,}665 MAG and 1 of 2{,}801 Prime queries correctly (\hone{} of 0.00, 0.00 and 0.04\%). Among all returned IDs (up to 20 per query), a gold ID appears for 1 Amazon and 4 Prime queries. The name probe obtains an \hone{} of 2.3\% on MAG (61 of 2{,}665) and 20.9\% on Prime (585 of 2{,}801). Table~\ref{tab:probe-strata} splits the queries by whether the name probe answers them. On the queries that the name probe answers incorrectly, \vfull{} achieves an \hone{} of 88.4 on MAG and 56.9 on Prime. On the 223 Prime queries of the test subset that the name probe answers incorrectly, the variants with access to the relations achieve between 55.6 and 63.2, while \vretr{} and \vnoedges{} achieve 28.7 and 28.3, respectively.

\begin{table}[t]
\centering
\footnotesize
\setlength{\tabcolsep}{3.5pt}
\begin{tabular}{@{}lrrrr@{}}
\toprule
 & \multicolumn{2}{c}{MAG} & \multicolumn{2}{c}{Prime} \\
\cmidrule(lr){2-3}\cmidrule(lr){4-5}
Name probe & right & wrong & right & wrong \\
\midrule
\multicolumn{5}{@{}l}{\emph{Full test set}} \\
Queries & 61 & 2{,}604 & 585 & 2{,}216 \\
\vfull{} & 67.2 & 88.4 & 89.7 & 56.9 \\
\midrule
\multicolumn{5}{@{}l}{\emph{Test subset}} \\
Queries & 4 & 262 & 57 & 223 \\
\vfull{} & 100.0 & 90.8 & 94.7 & 63.2 \\
\vtoolscode{} & 100.0 & 90.1 & 96.5 & 59.6 \\
\vtools{} & 100.0 & 87.4 & 94.7 & 59.2 \\
\vshell{} & 100.0 & 85.9 & 91.2 & 55.6 \\
\vretr{} & 100.0 & 47.7 & 87.7 & 28.7 \\
\vnoedges{} & 75.0 & 45.0 & 86.0 & 28.3 \\
\bottomrule
\end{tabular}
\caption{\hone{} of the Pi variants with GPT-6-Luna (all except \vone{}) on the queries that the name probe answers correctly (right) or incorrectly (wrong).}
\label{tab:probe-strata}
\end{table}

\section{Behavior Measures}
\label{app:behaviour}

\begin{table}[t]
\centering
\footnotesize
\setlength{\tabcolsep}{2.5pt}
\begin{tabular}{@{}lrrr@{}}
\toprule
 & \multicolumn{2}{c}{Pi} & Codex \\
\cmidrule(lr){2-3}\cmidrule(l){4-4}
 & Luna & Astra & Luna \\
\midrule
\emph{Write, run:} programs per query & 4.0 & 4.0 & 5.3 \\
\emph{Read:} programs with earlier IDs (\%) & 48.4 & 59.9 & 50.3 \\
\quad next turn runs Python (\%), after a & & & \\
\qquad truncated \texttt{rg} output & 66.8 & 76.1 & -- \\
\qquad untruncated \texttt{rg} output ${\geq}$4\,KB & 59.3 & 40.1 & -- \\
\emph{Revise:} revisions (\% of programs) & 36.5 & 33.7 & 36.7 \\
\quad next program revises it (\%), after a & & & \\
\qquad failed program & 88.6 & 92.2 & 85.1 \\
\qquad successful program & 49.7 & 38.0 & 42.7 \\
\bottomrule
\end{tabular}
\caption{Tool creation and tool use in the full-test trajectories of \vfull{} in Pi with GPT-6-Luna and GPT-6-Astra and in Codex with GPT-6-Luna. A program is one inline Python source, earlier IDs are node IDs shown in outputs of earlier turns, and truncated outputs were shortened by head-and-tail truncation in Pi. Codex truncates outputs by its own rule, and its two truncation cells show a dash. The measures are defined in this appendix.}
\label{tab:coevolution-app}
\end{table}

The measures in \S\ref{sec:analysis}, Figure~\ref{fig:coevolution}, Table~\ref{tab:coevolution-app} and the error analysis of Appendix~\ref{app:errors} are computed from the recorded trajectories of \vfull{} in Pi on the full test set (7,108 queries per model) and, for the Codex column of Table~\ref{tab:coevolution-app}, in Codex with GPT-6-Luna, and use each output as the model saw it. Pairs of values separated by a slash are for GPT-6-Luna and GPT-6-Astra, respectively.

\smallskip
\noindent \textbf{Programs, calls and turns.} A program is the inline Python code passed to the shell in one call, either as the argument of \texttt{python3 -c} or as a here-document read by \texttt{python3}. There are 28,582 / 28,726 programs, and 6,832 / 6,862 queries contain at least one. A turn is the set of calls issued in one model response. Because all calls in a turn are issued before the model sees any of their outputs, ``earlier'' and ``next'' refer to turns. The agent writes and runs all its code inline in its shell commands.

\smallskip
\noindent \textbf{Command intents and trajectory segments.} We assign every shell command to one of the nine categories that \citet{li2026beyond} use for DCI-Agent-Lite, with a rule-based classifier that parses the command's pipelines, and we apply the same classifier to the public DCI-Agent-Lite trajectories (789 queries) and to ours. A command that runs Python is \emph{Python Scripting}. Otherwise, a command with \texttt{wc} or with a search that counts matches is \emph{Aggregation/Counting}, a pipeline with two or more searches is \emph{Chain Search}, a search with a match limit or followed by \texttt{head} or \texttt{tail} is \emph{Search + Limit}, and any other search with \texttt{rg} or \texttt{grep} is \emph{Single Keyword Search}. A command with \texttt{find}, \texttt{rg -{}-files} or a search over a file listing is \emph{Locate File}, a single \texttt{cat} is \emph{Full Document Read}, \texttt{ls} or \texttt{pwd} is \emph{List Directory}, and all other commands are \emph{Other}. We merge the categories into Python (\emph{Python Scripting}), search (\emph{Search + Limit}, \emph{Chain Search} and \emph{Single Keyword Search}) and other (the remaining five). We compare the public trajectories of DCI-Agent-Lite~\citep{li2026beyond} with GPT-5.4-nano on 789 BrowseComp-Plus queries~\citep{chen2025browsecompplus}, \dkgi{} in Pi with GPT-6-Luna with the rg prompt and with our tool instruction on the same 450 validation queries (Appendix~\ref{app:validation}), and \vfull{} in Pi with GPT-6-Luna on the full test sets (7,108 queries, 60,032 shell commands). These counts cover shell commands, which exclude calls of Pi's \texttt{read} tool. Python commands make up 3.9\% of the shell commands of DCI-Agent-Lite, 1.5\% and 50.8\% with the rg prompt and our tool instruction, and 47.6\% for \vfull{}. Per dataset, 55.6\%, 41.5\% and 48.8\% of the shell commands of \vfull{} run Python on Amazon, MAG and Prime. To locate a call in its trajectory, we divide the $n$ calls of a query, including \texttt{read} calls, into five equal segments, so that the $i$-th call, counting from 0, lies in segment $\lfloor 5i/n\rfloor+1$. Over the calls of all queries, \texttt{read} and other calls make up the largest share in the first segment, \texttt{rg} calls in the second and Python calls in each later segment, with both models (Figure~\ref{fig:coevolution}a for GPT-6-Luna).

\smallskip
\noindent \textbf{Python commands and edge files.} A Python command is a shell call that runs \texttt{python3}. It refers to edge files if its code or arguments mention the edge directory or an edge file. With GPT-6-Luna, 41.1\% of the Python commands refer to edge files.

\smallskip
\noindent \textbf{Revisions.} The similarity of two programs is the ratio that Python's \texttt{difflib.SequenceMatcher} computes for their first 4,000 characters. Before comparing, we remove boilerplate lines: imports, comments, path constants, loops over node shards or over the lines of a file, \texttt{json.loads} of a line, and \texttt{try}/\texttt{except}-\texttt{continue} lines. A program is counted as revising an earlier program in its trajectory if its similarity to the most similar earlier program is at least 0.5. This holds for 36.5\% / 33.7\% of the programs. With each model, in 99.7\% of the revisions the program being revised was issued in an earlier turn, so its output was visible when the revision was written.

\smallskip
\noindent \textbf{IDs from earlier outputs.} A program contains an ID from an earlier output if its text contains a 4- to 8-digit integer that an output of an earlier turn showed as the ID of a node record, in an edge row, as a list element or at the start of a line. Years, multiples of 1,000 and numbers that appear in the prompt are excluded. This holds for 48.4\% / 59.9\% of the programs, and 83.1\% / 95.9\% of the queries with Python have such a program.

\smallskip
\noindent \textbf{Operation types.} We label each program by rules over its source code, and a program can have several labels. A \emph{full node scan} iterates over all node files. A \emph{one-hop traversal} reads an edge file, and a \emph{multi-hop traversal} filters an edge file by IDs that an earlier edge filter of the same program returned, or expands an adjacency index from a loop over seed nodes. Each program that reads an edge file gets one of these two traversal labels. The \emph{set operations} label covers intersections and differences of ID sets. \emph{Scoring or ranking} sorts candidates, keeps the top ones or accumulates scores or counts. A \emph{check of earlier IDs} contains an ID from an earlier output, as defined above, and looks it up in the node files, the edge files or the list of candidates. Two of these types, the one-hop traversal and the check of earlier IDs, have counterparts among the predefined graph tools (Table~\ref{tab:graph-tools}). \texttt{neighbors} performs a one-hop traversal, and \texttt{get\_node} looks up a known ID, as a check of earlier IDs does. The closest counterpart of a full node scan is \texttt{search}, which returns the $k\le30$ best BM25 matches for a query. The other three types can require several calls or code.

\smallskip
\noindent \textbf{Failures and the next program.} A program fails if it exits with a non-zero status, prints a Python traceback or syntax error, or times out. The 164 / 66 programs with an interpreter start error are excluded from both the failures and the successes. Of the programs that ran, 7.4\% / 8.5\% failed (2,089 / 2,426). The next program after a given program is the first program issued in a later turn. It is counted as revising the given program if their similarity, computed as above but with the boilerplate lines kept, is at least 0.5. Section~\ref{sec:analysis} and Table~\ref{tab:coevolution-app} use the 2,033 / 2,395 failed and 19,534 / 19,369 successful programs that have a next program. In 85.6\% / 91.1\% of these failures, the next program revises the failed one and runs successfully.

\smallskip
\noindent \textbf{Lineages and gold IDs in outputs.} We link each program to the most similar program issued before it in the same query if their similarity, computed with the boilerplate lines kept, is at least 0.5. The programs connected by these links form a \emph{lineage}, whose first and last versions are its earliest and latest programs. There are 7,282 / 7,621 lineages with at least two versions. A call shows a gold ID if its output that the model saw contains the ID of a gold node as a standalone number. For the first and last versions of the lineages, this holds for 40.0\% / 49.4\% of the first versions and 72.0\% / 67.9\% of the last versions. The test accepts the ID in any position of the output. Over the 5,080 queries that GPT-6-Luna answers correctly, the first call that shows a gold ID is a program in 68.5\%, an \texttt{rg} search in 29.7\% and another command in 1.8\%, while programs make up 38.2\% of all calls, \texttt{read} calls included. The error analysis of Appendix~\ref{app:errors} uses a stricter test that requires the node's record or its record header.

\smallskip
\noindent \textbf{Truncated outputs and the next turn.} An \texttt{rg} call is a shell call that runs \texttt{rg} or \texttt{grep} as a search, excluding file listings and calls that also run Python. We use the successful \texttt{rg} calls that are followed by another turn. An output is truncated if head-and-tail truncation (\S\ref{sec:outputs}) shortens it, and the outcome is whether the next turn runs a Python program. The model sees at least 14.8\,KB / 16.4\,KB of every truncated output, so we compare the 9,475 / 5,699 truncated outputs with the 1,577 / 2,169 untruncated outputs of at least 4\,KB. Because truncated and untruncated calls occur at different points of a trajectory, we compare them within each of the five trajectory segments, weight the segments by their number of truncated calls and drop segments with fewer than 20 calls on either side, which removes the first segment with GPT-6-Luna. With both models, the next turn runs Python more often after a truncated output than after an untruncated one of at least 4\,KB (66.8\% vs.\ 59.3\% with GPT-6-Luna and 76.1\% vs.\ 40.1\% with GPT-6-Astra, Table~\ref{tab:coevolution-app}).

\smallskip
\noindent \textbf{Text-constraint tags.} GPT-6-Luna, run with reasoning effort \emph{high}, labels each query from two inputs, the query text and the names of the node and relation types. A query has a text constraint if checking one of its requirements needs a node's own text. The LLM annotator tags 96\% of the Amazon and 57\% of the Prime queries as having an additional text constraint. We treat all MAG queries as text-constrained based on their topic descriptions. On Amazon, \hone{} is 8.1 points higher on text-constrained queries than on the 60 relation-only queries, and on Prime it is 2.1 points lower.

\section{Error Analysis}
\label{app:errors}

We assign every query that \vfull{} in Pi answers incorrectly on the full test sets (7,108 queries per backbone) to one of four categories, using the outputs as the model saw them. A query is an \emph{empty answer} if the run ends in a failure, in output other than the requested JSON object or in an empty list. Otherwise, it is a \emph{ranking error} if a gold node is returned at rank 2 or lower, a \emph{selection error} if an output that the model saw showed a gold record and the returned list misses every gold node, and a \emph{search miss} in all remaining cases. An output shows a gold record if it contains the node's record (\texttt{\qq{}node\_id\qq{}:}~\dkgiph{id}) or the ID alone in square brackets (\texttt{[}\dkgiph{id}\texttt{]}), as in a record header. IDs in other positions, for example in edge rows, are excluded.

\begin{table}[t]
\centering
\footnotesize
\setlength{\tabcolsep}{3pt}
\begin{tabular}{@{}lrrrrr@{}}
\toprule
 & & \multicolumn{4}{c}{Share of errors (\%)} \\
\cmidrule(l){3-6}
 & & Ranking & Selection & Search & Empty \\
Dataset & \hone{} & error & error & miss & answer \\
\midrule
\multicolumn{6}{@{}l}{\emph{GPT-6-Luna}} \\
Amazon & 57.8 & 58 & 1 & 41 & 0 \\
MAG & 88.0 & 46 & 26 & 23 & 6 \\
Prime & 63.8 & 55 & 7 & 37 & 0 \\
Overall & 71.5 & 55 & 8 & 36 & 1 \\
\midrule
\multicolumn{6}{@{}l}{\emph{GPT-6-Astra}} \\
Amazon & 79.4 & 72 & 2 & 26 & 0 \\
MAG & 91.8 & 49 & 38 & 10 & 4 \\
Prime & 74.5 & 58 & 14 & 28 & 0 \\
Overall & 82.1 & 60 & 15 & 24 & 1 \\
\bottomrule
\end{tabular}
\caption{Error categories of \vfull{} in Pi on the full test sets (7,108 queries) with GPT-6-Luna and GPT-6-Astra. The four categories divide the incorrectly answered queries, and each share is a percentage of these queries, rounded to a whole number.}
\label{tab:errors}
\end{table}

Ranking errors are the largest category on every dataset with both backbones and make up 55\% of the errors overall with GPT-6-Luna and 60\% with GPT-6-Astra (Table~\ref{tab:errors}). In these queries, the agent has found a gold node and returned it below the first position, so a gold node is found in more queries than \hone{} shows. With the stronger GPT-6-Astra, search misses make up 24\% of the errors overall, compared with 36\% with GPT-6-Luna, and empty answers make up 1\% of the errors overall with both backbones.

\section{Case Studies}
\label{app:cases}

We compare \vfull{} with \vtools{} on three requests from the 710-query test subset, one from each dataset. On each of them, \vfull{} ranks a gold node first and \vtools{} ranks another node first. Table~\ref{tab:case-outcomes} lists the outcome and the number of calls of every configuration except \vone{} on these requests. Each case gives the request and its gold answers, an excerpt of each trajectory with GPT-6-Luna, and a comparison of the operation that the agent writes with the calls that the graph tools of Table~\ref{tab:graph-tools} provide.

The request and its gold answers are shown in a gray box. In the excerpts, \texttt{...} marks omitted code or output, a bold \texttt{[k]} gives the index of the call, and \texttt{->} introduces the output that the model saw, shown in gray. A \vtools{} call is shown as the tool name followed by its arguments. A line too long for the box continues after the mark \dkgiwrapmark{}, and every other line begins a new line of the code or output and keeps its original indentation. A green box title marks a configuration that ranks a gold node first, and a red title marks one that ranks another node first. IDs are benchmark node IDs.

\begin{table*}[t]
\centering
\footnotesize
\setlength{\tabcolsep}{5pt}
\begin{tabular}{@{}lrrrrrr@{}}
\toprule
 & \multicolumn{2}{c}{Case 1} & \multicolumn{2}{c}{Case 2} & \multicolumn{2}{c}{Case 3} \\
 & \multicolumn{2}{c}{MAG} & \multicolumn{2}{c}{Prime} & \multicolumn{2}{c}{Amazon} \\
\cmidrule(lr){2-3}\cmidrule(lr){4-5}\cmidrule(l){6-7}
Configuration & Rank & Calls & Rank & Calls & Rank & Calls \\
\midrule
\vfull{}, GPT-6-Luna & 1 & 21 & 1 & 13 & 1 & 13 \\
\vfull{}, GPT-6-Astra & 1 & 10 & 1 & 5 & 1 & 10 \\
\vcodex{} & 1 & 12 & 1 & 9 & 1 & 11 \\
\vtoolscode{} & 1 & 11 & 1 & 9 & 1 & 4 \\
\vtools{} & -- & 38 & 3 & 36 & 2 & 23 \\
\vshell{} & -- & 17 & -- & 39 & -- & 23 \\
\vretr{} & -- & 15 & -- & 19 & 2 & 16 \\
\vnoedges{} & -- & 14 & -- & 12 & -- & 7 \\
\bottomrule
\end{tabular}
\caption{Outcome of each configuration on the three case requests. Rank: rank of the first gold node in the returned list (rank 1 is correct under \hone{}). --: every returned node is incorrect. Calls: the agent's actions $a_t$ of \S\ref{sec:method} (for \vtoolscode{}, \texttt{bash} calls, each of which may call several graph tools, and for \vcodex{}, command executions). All rows except the second use GPT-6-Luna.}
\label{tab:case-outcomes}
\end{table*}

\subsection{Case 1 (MAG): A Program Grows into a Multi-Hop Traversal}
\label{app:case1}

\begin{dkgiquery}{STaRK-MAG request}
\textbf{Request:} Papers on data quality enhancement algorithms affiliated with the University of the Balearic Islands.

\textbf{Gold:} one paper, 1778273 (\emph{Effects of data quality vetoes on a search for compact binary coalescences \dots}).
\end{dkgiquery}

The KG links the institution to 66 authors and these authors to 236 papers. Of the 86 papers with the field of study \emph{Data quality}, one is among these 236.

\smallskip
\noindent \textbf{\vtools{}.} The agent finds the institution and the field \emph{Data quality} with \texttt{search}. Its \texttt{neighbors} call on the institution lists 20 of the 66 affiliated authors (call 6), and two of these 20 authors are co-authors of the gold paper. Its \texttt{neighbors} call on the field lists 20 of the 86 papers in the field (call 11). The agent then lists the papers of four other authors from the list, one call per author (calls 27--30), and the authors of nine papers found by search, one call per paper. The gold paper is absent from all 38 outputs.

\begin{dkgicase*}[breakable]{Case 1 (MAG): \vtools{}, GPT-6-Luna}
@<[6]>@  neighbors {"node_id": 1106133, "relation":
     ~>"author___affiliated_with___institution",
     ~>"node_type": "author", "k": 20}
     @|-> Search around [1106133] University of the
      ~>Balearic Islands (...) found 66 results:
        0. [6817] Denis Pollney (author) ...
        ...
        (Showing 20 of 66 results. You may want to
      ~>refine your search.)|@
@<[11]>@ neighbors {"node_id": 1116111, "relation":
     ~>"paper___has_topic___field_of_study",
     ~>"node_type": "paper", "k": 20}
     @|-> Search around [1116111] Data quality (...)
      ~>found 86 results:
        ...
        (Showing 20 of 86 results. ...)|@
@<[27]>@ neighbors {"node_id": 26073, "relation":
     ~>"author___writes___paper", "node_type":
     ~>"paper", "k": 20}
     @|-> Search around [26073] Ramon Pujol-Nadal (...)
      ~>found 1 result:
        0. -author___writes___paper-> [1682034]
      ~>Influence of thermal losses ...|@
\end{dkgicase*}

\smallskip
\noindent \textbf{\vfull{}.} The first program (call 11) has a silent bug. Each edge line still ends in its newline character, so the membership test fails on every line, and the program prints the two file names. After printing the file headers (call 12), the agent revises the program (call 13). The next program is a two-hop traversal from the institution through its 66 authors to their 236 papers, and it reads the node shards that hold these papers (call 14). Call 15 adds the topic relation and an intersection, with the field ID shown by the \texttt{rg} search of call 9, and the intersection holds one paper, the gold paper. Call 19 rewrites the same traversal more compactly to rank the candidates.

\begin{dkgicase}[breakable]{Case 1 (MAG): \vfull{}, GPT-6-Luna}
@<[4]>@  rg -n -i 'University of the Balearic ...' nodes
     @|-> ...{"node_id": 1106133, "node_type":
      ~>"institution", ... "DisplayName":
      ~>"University of the Balearic Islands", ...|@
@<[9]>@  rg -n -i 'data quality|...' nodes/0110.jsonl
     ~>nodes/0*.jsonl
     @|-> ...{"node_id": 1116111, "node_type":
      ~>"field_of_study", ... "DisplayName":
      ~>"Data quality", ...
        [... harness output limit: ... of 2107
      ~>lines ...]|@
@<[11]>@ p=1106133
     for f in ['edges/00.tsv','edges/03.tsv']:
      ...
        if str(p) in line.split('\t'):
     @|-> edges/00.tsv
        edges/03.tsv|@
@<[13]>@ for f in ['edges/00.tsv','edges/03.tsv',...]:
      ...
        a,b=map(int,line.split())
        if a==1106133 or b==1106133:
     @|-> edges/00.tsv 66 [(6817, 1106133), ...|@
@<[14]>@ ... int(row[1])==inst: authors.add(int(row[0]))
     ... in authors: papers.add(int(row[1]))
     for sh in sorted({x//10000 for x in papers}):
     @|-> authors 66 [6817, 16943, 18306, ...]
        papers 236|@
@<[15]>@ fields={1116111}
     ... int(r[1]) in fields:FP.add(int(r[0]))
     print('P affiliated',len(P),...,len(P&FP))
     @|-> P affiliated 236 field data quality papers
      ~>86 intersect 1
        1778273 Effects of data quality vetoes ...|@
@<[19]>@ A={int(r[0]) for r in ... int(r[1])==1106133}
     P={int(r[1]) for r in ... int(r[0]) in A}
     ... score=sum(bool(re.search(p,s)) for p in
     ~>[r'data quality',r'quality of the data',...])
\end{dkgicase}

\smallskip
\noindent \textbf{Comparison.} The agent writes the request as a traversal over three relations, from the institution to its authors, from the authors to their papers and from the field to its papers, followed by an intersection of the two sets of papers. With the predefined graph tools, each hop of this traversal is a \texttt{neighbors} call from a single node that lists up to 20 neighbors and the total number found, and \vtools{} makes one such call for each author or paper that it examines.

\subsection{Case 2 (Prime): A Program as Set Algebra over Two Relations}
\label{app:case2}

\begin{dkgiquery}{STaRK-Prime request}
\textbf{Request:} Which genes or proteins are expressed exclusively in the pericardium and not in either the dorsal or ventral regions of the thalamus?

\textbf{Gold:} ADH1B (10833).
\end{dkgiquery}

The pericardium has \emph{expression present} edges to 13 genes. Nine of them also have an \emph{expression present} edge to \emph{dorsal plus ventral thalamus}, and of the 13, ADH1B alone has an \emph{expression absent} edge to this node. In the edge files, relation 16 is \emph{expression present} and relation 17 \emph{expression absent}.

\smallskip
\noindent \textbf{\vtools{}.} The agent lists the 13 pericardium genes with one \texttt{neighbors} call (call 5). Its requests for \emph{expression absent} edges go to the pericardium and to the nodes \emph{dorsal thalamus} and \emph{ventral thalamus}, and all three return empty results (calls 8, 10 and 11). It then tests the genes one at a time, with a \texttt{get\_node} call for each gene (calls 12--24) and, for 12 of them, a \texttt{neighbors} call on \emph{dorsal plus ventral thalamus} that is restricted to \emph{expression present} and to the gene's name (calls 25--36). It returns the four genes that this test leaves, PDLIM3, TCAP, ADH1B and ITLN1, in this order.

\begin{dkgicase*}[breakable]{Case 2 (Prime): \vtools{}, GPT-6-Luna}
@<[5]>@  neighbors {"node_id": 64835, "relation":
     ~>"expression present", "node_type":
     ~>"gene/protein", "k": 20}
     @|-> Search around [64835] pericardium (...) found
      ~>13 results:
        0. -expression present-> [61] CALM1 ...
        ...
        10. -expression present-> [10833] ADH1B ...
        ...|@
@<[10]>@ neighbors {"node_id": 64341, "relation":
     ~>"expression absent", "node_type":
     ~>"gene/protein", "k": 20}
     @|-> Search around [64341] ventral thalamus (...)
      ~>found no results. ...|@
@<[34]>@ neighbors {"node_id": 64338, "relation":
     ~>"expression present", "node_type":
     ~>"gene/protein", "query": "ADH1B", "k": 5}
     @|-> Search around [64338] dorsal plus ventral
      ~>thalamus (...) found no results. ...|@
@<answer:>@ [1403, 2773, 10833, 13398]
\end{dkgicase*}

\smallskip
\noindent \textbf{\vfull{}.} After a scan of the node records for the anatomy nodes of the request (call 5), the agent writes one program over both relations (call 12). It intersects the 13 pericardium genes, whose IDs it copies from the output of call 11, with the \emph{expression present} and \emph{expression absent} neighbors of seven thalamus nodes. The output shows nine genes with an \emph{expression present} edge to \emph{dorsal plus ventral thalamus} and ADH1B with an \emph{expression absent} edge to it. The agent returns the four remaining genes and ranks ADH1B first.

\begin{dkgicase}[breakable]{Case 2 (Prime): \vfull{}, GPT-6-Luna}
@<[5]>@  ... if x.get("node_type")=="anatomy" and
     ~>any(t in s for t in terms): print(...)
     @|-> ...
        64338 True {... 'name': 'dorsal plus
      ~>ventral thalamus', ...}
        64341 True {... 'name': 'ventral thalamus',
      ~>...}
        ...
        64835 True {... 'name': 'pericardium', ...}
        ...|@
@<[12]>@ p={61,652,704,1403,...,10833,13398,13628};
     for r in (16,17):
      for anat in [64338,64341,66732,71295,...]:
       q=set()
       for e in csv.DictReader(
       ~>open(f"edges/{r:02}.tsv"),delimiter="\t"):
        ...
        if anat in (a,b): q.add(b if a==anat else a)
       print(r,anat,"overlap",sorted(p&q))
     @|-> 16 64338 overlap [61, 652, 704, 1532, 2500,
         ~>3225, 3387, 9724, 13628]
        ...
        17 64338 overlap [10833]|@
@<answer:>@ [10833, 1403, 2773, 13398]
\end{dkgicase}

\smallskip
\noindent \textbf{Comparison.} The agent's program performs set algebra over two relations. In one run, it intersects a set of genes with the neighbors of seven anatomy nodes under both relations, and the agent takes the difference from the printed intersections. With the predefined graph tools, each test covers one node, one relation and, through \texttt{query}, one gene name, and \vtools{} spends 25 calls on the 13 genes.

\subsection{Case 3 (Amazon): An Error, Its Repair and Four Narrower Filters}
\label{app:case3}

\begin{dkgiquery}{STaRK-Amazon request}
\textbf{Request:} What are some officially licensed NCAA brand metal tags for a hardcore fan looking to enhance their collection and showcase their team spirit?

\textbf{Gold:} 18 products titled \emph{Licensed NCAA $\langle$team$\rangle$ Metal Tag~\dots}
\end{dkgiquery}

In the Amazon records, the reviews precede the title, so a record line printed by \texttt{rg} consists mostly of review text.

\smallskip
\noindent \textbf{\vtools{}.} The first search returns 30 products, with \emph{Rico Industries NCAA Unisex NCAA \#1 Fan Metal Auto Tag} first and the 18 gold products in places 2 to 19 (call 2). The agent opens 20 products from its searches, one \texttt{get\_node} call each, and returns them with the Rico tag first and the 18 gold products at ranks 2 to 19.

\begin{dkgicase*}[breakable]{Case 3 (Amazon): \vtools{}, GPT-6-Luna}
@<[2]>@  search {"query": "officially licensed NCAA
     ~>brand metal tag hardcore fan collection team
     ~>spirit NCAA metal tags", "k": 30}
     @|-> Found 30 nodes for query: ...
        [831452] Rico Industries NCAA Unisex NCAA #1
      ~>Fan Metal Auto Tag (product) - Score: 77.007
        ...
        [631548] Licensed NCAA KU Kansas Jayhawks #1 Fan
      ~>Logo Metal Tag, ... (product) - Score: 72.931
        ...|@
@<[3]>@  get_node {"node_id": 831452}
     @|-> {"node_id":831452,"node_type":"product",
      ~>"is_candidate":true,"attributes":{"review":
      ~>[{"reviewerID": ...|@
@<answer:>@ [831452, 631548, 631550, 631544, 631551, ...]
\end{dkgicase*}

\smallskip
\noindent \textbf{\vfull{}.} The first program fails because \texttt{\textbackslash n} is written literally inside the single-quoted argument of \texttt{python3 -c} (call 6). The next call passes the same code (similarity 0.99) inside \texttt{\$'...'} quotes, and this program runs (call 7). Each of the next four programs (calls 8, 9, 10 and 12) follows a truncated output that lists unrelated products (``Tagless collars'', key chains, domino sets in metal tins) and narrows the filter. The last program looks up the titles of the 27 IDs that the previous output listed (call 13), and the agent returns the 18 gold products at ranks 1 to 18.

\begin{dkgicase}{Case 3 (Amazon): \vfull{}, GPT-6-Luna}
@<[6]>@  python3 -c 'import json,glob,re; ...;\nfor f
     ~>in glob.glob("nodes/*.jsonl"):\n ...'
     @|-> ... SyntaxError: unexpected character
      ~>after line continuation character|@
@<[7]>@  python3 -c \$'import json,glob,re\n...'
     @|-> 330040 NIKE Colorado Buffaloes Infant ...
      ~>| Nike | 100% Cotton ... Tagless collars ...
        [... harness output limit: ... of 9824
      ~>lines ...]|@
@<[8]>@  ... re.search(r"tag",s,re.I) and
     ~>re.search(r"NCAA|colleg|university|...",s,re.I)
@<[9]>@  ... re.search(r"metal",t,re.I) and ...
     @|-> 330440 Florida State Seminoles Chrome Metal
      ~>Key Chain ...|@
@<[10]>@ ... re.search(r"NCAA|...|Hurricanes",t,re.I) and
     ~>re.search(r"metal",t+" "+s,re.I) and
     ~>re.search(r"tag|license plate",t+" "+s,re.I)
     @|-> 331243 NCAA LSU Tigers Full Color License
      ~>Plate Frame, ...
        335385 NCAA Kansas Jayhawks Domino Set in
      ~>Metal Gift Tin ...|@
@<[12]>@ ... re.search(r"NCAA",t+" "+s,re.I) and
     ~>re.search(r"dog tag|metal tag|...",t+" "+s,re.I)
     @|-> ... 631540 Licensed NCAA Ole Miss Metal Tag,
      ~>(Proudly Made In The USA), ...|@
@<[13]>@ ids={631540,631541,631544,...,906791}
     ... if x["node_id"] in ids: print(...)
\end{dkgicase}

\smallskip
\noindent \textbf{Comparison.} The agent's tool is a filter over the title, features and description of every product record, and the agent narrows it four times after truncated outputs. The predefined graph tools provide a BM25 search over node names and summaries and \texttt{get\_node}, which returns one complete record per call, with the reviews before the title.

\end{document}